\documentclass[10pt,journal,cspaper,compsoc]{IEEEtran}

\usepackage{epsfig}
\usepackage{graphicx}
\usepackage{amsmath}
\usepackage{amssymb}
\usepackage{multirow}
\usepackage{bbding}
\usepackage{pifont}
\usepackage{caption}
\usepackage[bookmarks=false,colorlinks=true,linkcolor=black,citecolor=black,filecolor=black,urlcolor=black]{hyperref}

\newcommand{\conv}[1]{$\left[\begin{array}{ll} \text{1}\times \text{1} \text{ conv}\\ \text{3}\times \text{3} \text{ conv} \end{array}\right] \times \text{#1}$}

\newcommand{\cross}[1]{#1 $\times$ #1}

\begin{document}

%
\title{Object Detection from Scratch with Deep Supervision}

\author{Zhiqiang~Shen, Zhuang~Liu, Jianguo Li, Yu-Gang~Jiang,  Yurong~Chen and Xiangyang~Xue
\IEEEcompsocitemizethanks{
\IEEEcompsocthanksitem Zhiqiang~Shen, Yu-Gang~Jiang and Xiangyang~Xue are with the School
of Computer Science, Fudan University, Shanghai, China. Zhiqiang Shen is now a visiting scholar at UIUC. Yu-Gang Jiang is also with Jilian Technology Group (Video++), Shanghai, China. This work was done when Zhiqiang Shen and Zhuang Liu were interns at Intel Labs China. 
E-mail: zhiqiangshen0214@gmail.com, \{ygj, xyxue\}@fudan.edu.cn.
\IEEEcompsocthanksitem Zhuang~Liu was with the Institute 
    for Interdisciplinary Information Sciences, Tsinghua University, Beijing, China. Now he is a PhD student at UC Berkeley. 
	E-mail: liuzhuangthu@gmail.com.
\IEEEcompsocthanksitem Jianguo Li and Yurong~Chen are with Intel Labs China,
   Beijing, China. E-mail: \{jianguo.li, yurong.chen\}@intel.com.}
}

\markboth{}%
	{Shell \MakeLowercase{\textit{et al.}}: Bare Demo of IEEEtran.cls for Computer Society Journals}

\IEEEcompsoctitleabstractindextext{%
\begin{abstract}

In this paper, we propose \textbf{D}eeply \textbf{S}upervised \textbf{O}bject \textbf{D}etectors (DSOD), an object detection framework that can be trained from scratch. Recent advances in object detection heavily depend on the off-the-shelf models pre-trained on large-scale classification datasets like ImageNet and OpenImage. However, one problem is that adopting pre-trained models from classification to detection task may incur learning bias due to the different objective function and diverse distributions of object categories. Techniques like fine-tuning on detection task could alleviate this issue to some extent but are still not fundamental. Furthermore, transferring these pre-trained models across discrepant domains will be more difficult (e.g., from RGB to depth images). Thus, a better solution to handle these critical problems is to train object detectors from scratch, which motivates our proposed method. Previous efforts on this direction mainly failed by reasons of the limited training data and naive backbone network structures for object detection. In DSOD, we contribute a set of design principles for learning object detectors from scratch. One of the key principles is the {\em deep supervision}, enabled by layer-wise dense connections in both backbone networks and prediction layers, plays a critical role in learning good detectors from scratch. After involving several other principles, we build our DSOD based on the single-shot detection framework (SSD). We evaluate our method on PASCAL VOC 2007, 2012 and COCO datasets. DSOD achieves consistently better results than the state-of-the-art methods with much more compact models. Specifically, DSOD outperforms baseline method SSD on all three benchmarks, while requiring only 1/2 parameters. We also observe that DSOD can achieve comparable/slightly better results than Mask RCNN~\cite{he2017mask} + FPN~\cite{lin2016feature} (under similar input size) with only 1/3 parameters, using no extra data or pre-trained models. 

\end{abstract}

}

\maketitle


\IEEEpeerreviewmaketitle

\section{Introduction}\label{sec:introduction}

\IEEEPARstart{G}{eneric} object detection is the task that we aim to localize various objects in a natural image automatically. This task has been heavily studied due to its wide applications in surveillance, autonomous driving, intelligent security, etc.
In the recent years, with the progress of more and more innovative and powerful Convolutional Neural Networks (CNNs) based object detection systems have been proposed, the object detection problem has been one of the fastest moving areas in computer vision.

\begin{figure*}[]
	\centering
	\includegraphics[width=0.90\textwidth]{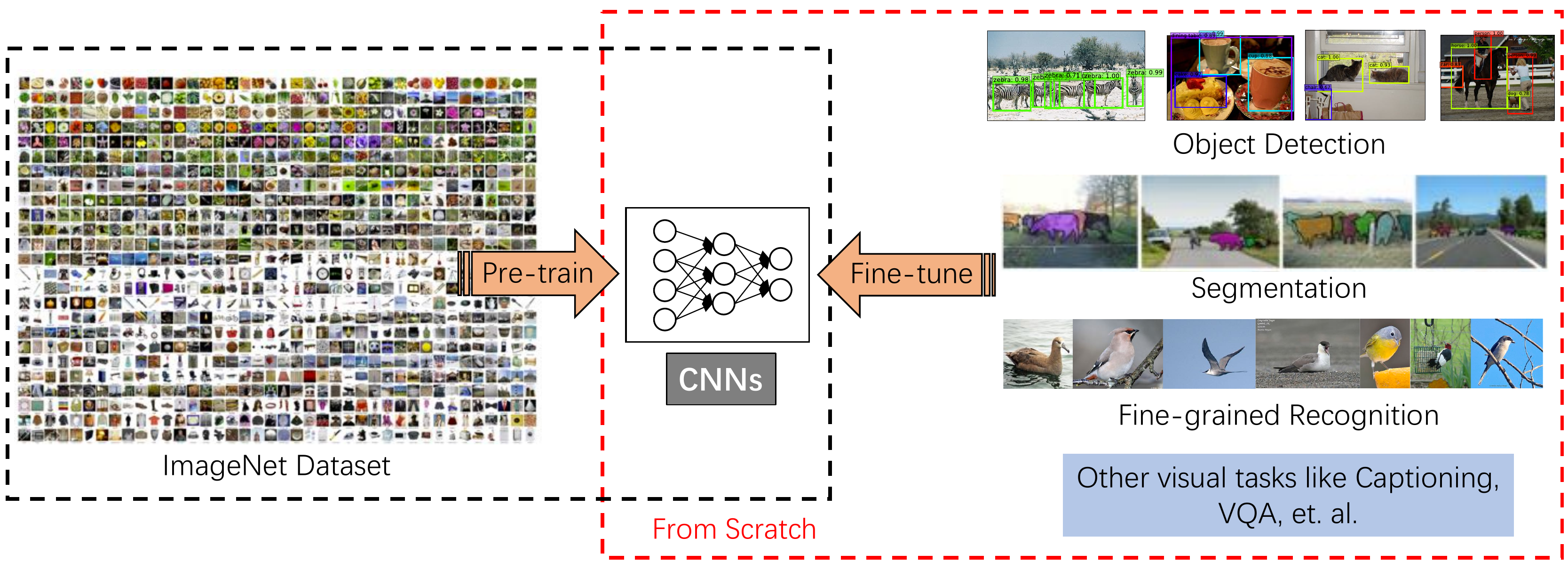}
	\vspace{-0.1in}
	\caption{Illustration of \textbf{Training Models from Scratch}. The black dashed box (left) denotes we pre-train models on large-scale classification dataset like ImageNet~\cite{deng2009imagenet}. The red dashed box (right) denotes we train models on target dataset directly. In this paper, we focus on the object detection task without using the pre-trained models.}
	\label{scratch}
\end{figure*}

To achieve desired performance, the common practice in advanced object detection systems is to fine-tune models
pre-trained on ImageNet~\cite{deng2009imagenet}. This fine-tuning process can be viewed as transfer learning~\cite{oquab2014learning,cui2018transfer}. Specifically, as is shown in Fig.~\ref{scratch}, researchers usually train CNN models on large-scale classification datasets like ImageNet~\cite{deng2009imagenet} first, then fine-tune the models on target tasks, such as 
object detection~\cite{girshick2014rich,girshick2015fast,ren2015faster,li2016r,he2017mask,liu2016ssd,redmon2016you,lin2017focal,kong2016hypernet,bell2016inside,kong2017ron,peng2018megdet,singh2018analysis,hu2018relation,cai2018cascade,bosquetstdnet,xu2018deep,wang2018pelee}, image segmentation~\cite{long2015fully,hariharan2015hypercolumns,chen2014semantic,yu2015multi}, fine-grained recognition~\cite{zhang2014part,lin2015bilinear,wang2015multiple,krause2015fine}, captioning~\cite{xu2015show,donahue2015long,fang2015captions,shen2017weakly,krishna2017dense,johnson2016densecap}, etc. {\em Learning from scratch} means we directly train models on these target tasks without involving any other additional data or extra fine-tuning processes. Empirically,
fine-tuning from pre-trained models has at least two advantages.
First, there are numerous state-of-the-art pre-trained CNN models publicly available.
It is convenient for researchers to reuse the learned parameters in their own domain-specific tasks.
Second, fine-tuning on pre-trained models can quickly convergence to a final state and requires less instance-level annotated training data than basic classification task.

However, the critical limitations are also obvious when adopting the pre-trained models for object detection:
(I) {\em{Limited design space on network structures}}.
Existing object detectors directly adopt the pre-trained networks,
and as a consequence, there is little flexibility to control/adjust the detailed network structures, even for small changes of network design.
Furthermore, the pre-trained models are mostly from large-scale classification task, which are usually very heavy (containing a huge number of parameters) and are not suitable for some specific scenarios. The heavy network structures will bound the requirement of computing resources.
(II) {\em{Learning/optimization bias}}.
Since there are some differences in both the objective functions and the category distributions between classification and detection tasks,
these differences may lead to different searching/optimization spaces.
Therefore, learning may be biased towards a local minimum when all parameters are initialized from classification pre-trained models, which is not the best for target detection task.
(III) {\em{Domain mismatch}}. As is well-known, fine-tuning can mitigate the gap between different target category distribution.
However, it is still a severe problem when the source domain (e.g., ImageNet) has a huge mismatch to the target domain such as depth images, medical images, etc~\cite{gupta2016cross}.

Therefore, our work is motivated by the following two questions.
First, is it possible to train object detection networks from scratch directly without the pre-trained models?
Second, if the first answer is positive,
are there any principles to design a resource efficient network structure for object detection, meanwhile keeping high detection accuracy?
To meet this goal, we propose deeply supervised objection detectors (DSOD), a simple yet efficient framework that can learn object detectors from scratch.
DSOD is fairly flexible, we can tailor various network structures for different computing platforms such as servers, desktop, mobile and even embedded devices.

\begin{figure*}[t]
	\centering
	\includegraphics[width=0.8\textwidth]{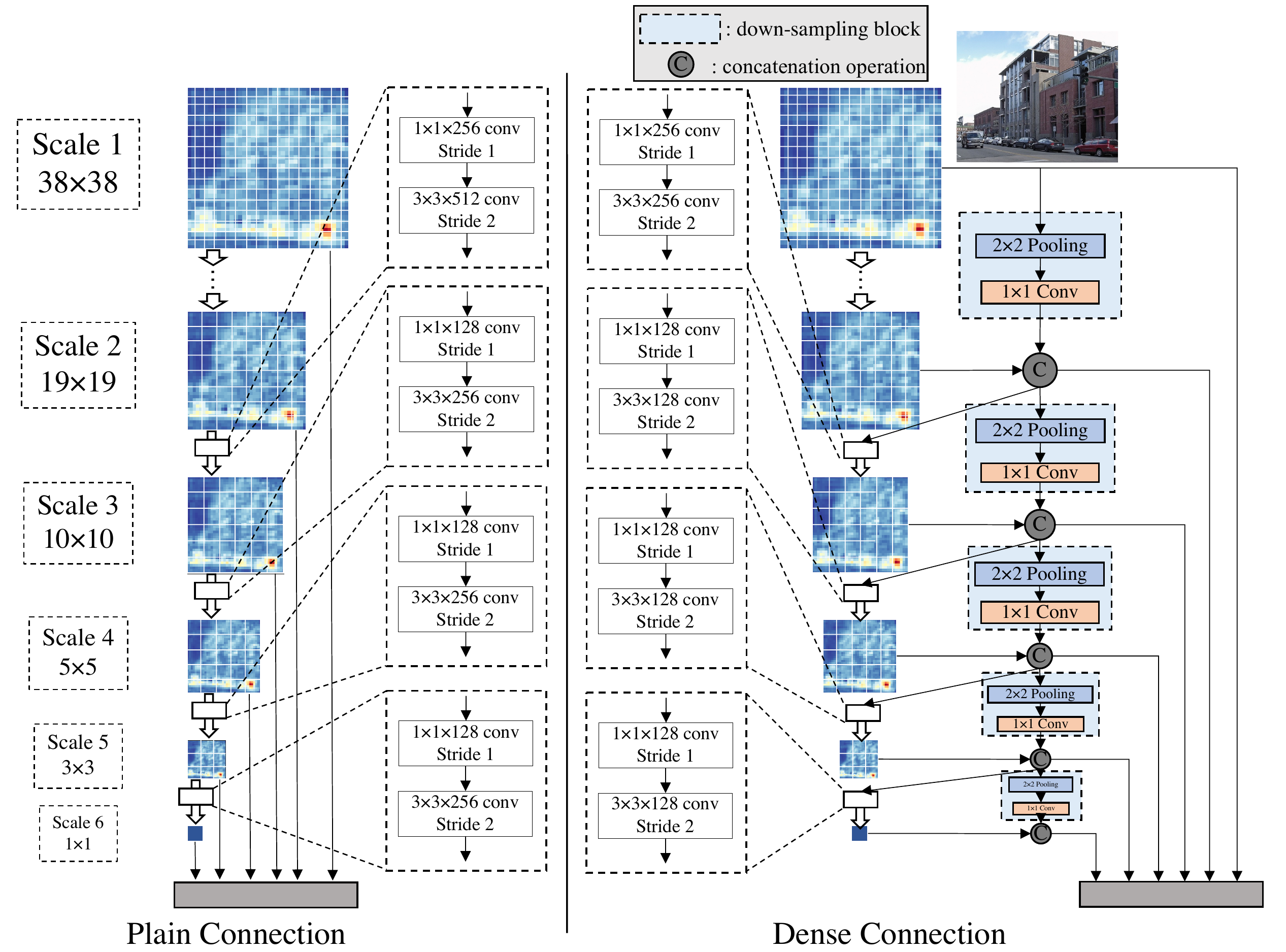}
	\vspace{-0.12in}
	\caption{DSOD prediction layers with plain and dense structures (for 300$\times$300 input). 
		The plain structure is introduced by SSD~\cite{liu2016ssd} and dense structure is ours. 
		See Section~\ref{model} for more details.}
	\label{dense_connection}
\end{figure*}

We contribute a set of principles for designing DSOD.
One key point is the {\em{deeply supervised}} structure, which is motivated by the recent work of~\cite{lee2015deeply,xie2015hed}. In~\cite{xie2015hed}, Xie {\em{et al.}} proposed a holistically-nested structure for edge detection, which included the side-output layers in each conv-stage of base network for explicit deep supervision.  
Instead of using the multiple cut-in loss signals with side-output layers, our method adopts {\em{deep supervision}} implicitly through the layer-wise dense connections proposed in DenseNet~\cite{huang2016densely}.
Dense structures are not only adopted in the backbone sub-network, but also used in the front-end multi-scale prediction layers.
Fig.~\ref{dense_connection} illustrates the structure comparison in front-end prediction layers between baseline SSD and our DSOD.
The fusion and reuse of multi-resolution prediction-maps help keep or even improve the final accuracy, while reducing model parameters to some extent. As shown in Fig.~\ref{ds}, we further adopted dense connections between different blocks to enhance the deeply supervised signals during network training.

Furthermore, we revisited the pre-activation BN-Conv-ReLU of backbone networks for our DSOD framework. We observe that post-activation  (Conv-BN-ReLU) order can obtain about 0.6\% mAP improvement on VOC 07, meanwhile, requiring slightly fewer parameters compared with original order in DSOD. In order to further enhance the deep supervision purpose when training from scratch, especially for some plain backbones like VGGNet, we also propose a complementary structure named deep-scale supervision module (DSS) as DSOD v2. More details are given in the following sections.
%
Now, we summarize our main contributions of this paper as follows:
\begin{itemize}
	\addtolength{\itemsep}{-0in}
	\item[(1)] To the best of our knowledge, DSOD is the first framework that can train object detectors from scratch with promising performance.
	\item[(2)] We introduce and validate a set of principles to design efficient object detection networks from scratch through step-by-step ablation studies.
	\item[(3)] We show that DSOD can achieve comparable performance with state-of-the-arts on three standard benchmarks (PASCAL VOC 2007, 2012 and MS COCO datasets), meanwhile, has real-time processing speed and more compact models.
\end{itemize}

A preliminary version of this manuscript~\cite{Shen2017DSOD} has been published on a previous conference. In this version, we made some design changes in backbone network (e.g., replacing pre-activation in BN-ReLU-Conv with the post-activation Conv-BN-ReLU manner) and included a new module (named deep-scale supervision) to make DSOD better (Section~\ref{32}). We also included more details, analysis and extra comparison experiments with state-of-the-art two-stage detectors like FPN and Mask RCNN and the factors of training them from scratch (Section~\ref{48},~\ref{49} and \ref{discussion}). The proposed DSOD framework has also been adopted and generalized to further improve the performance under the setting of learning object detectors from scratch such as GRP-DSOD~\cite{shen2017learning}, Tiny-DSOD~\cite{li2018tiny}, etc.

\renewcommand{\arraystretch}{1.1}
\begin{table*}[]
	\centering
    \caption{DSOD architecture (growth rate $k$ = 48 in each dense block).}
	\label{base_network}
	\resizebox{0.80\textwidth}{!}{%
		\begin{tabular}{c|c|c|c}
			\hline
			\multicolumn{2}{c|}{Layers}                                & Output Size (Input 3$\times$\cross{300}) & DSOD                                                               \\ \hline
			\multirow{4}{*}{Stem}             & Convolution             &64$\times$150$\times$150  & 3$\times$3 conv, stride 2                                                     \\ \cline{2-4}
			& Convolution             & 64$\times$150$\times$150  & 3$\times$3 conv, stride 1                                                     \\ \cline{2-4}
			& Convolution             & 128$\times$150$\times$150 & 3$\times$3 conv, stride 1                                                     \\ \cline{2-4}
			& Pooling                 & 128$\times$75$\times$75   & 2$\times$2 max pool, stride 2                                                \\ \hline
			\multicolumn{2}{c|}{\begin{tabular}[c]{@{}c@{}}Dense Block\\ (1)\end{tabular}}                       & 416$\times$75$\times$75   & \conv{6} \\ \hline
			\multicolumn{2}{c|}{\multirow{2}{*}{\begin{tabular}[c]{@{}c@{}}Transition Layer\\ (1)\end{tabular}}} & 416$\times$75$\times$75   & 1$\times$1 conv                                                              \\ \cline{3-4}
			\multicolumn{2}{c|}{}                                      & 416$\times$38$\times$38   & 2$\times$2 max pool, stride 2                                                \\ \hline
			\multicolumn{2}{c|}{\begin{tabular}[c]{@{}c@{}}Dense Block\\ (2)\end{tabular}}                       & 800$\times$38$\times$38   &                     \conv{8}                                              \\ \hline
			\multicolumn{2}{c|}{\multirow{2}{*}{\begin{tabular}[c]{@{}c@{}}Transition Layer\\ (2)\end{tabular}}} & 800$\times$38$\times$38   & 1$\times$1 conv                                                              \\ \cline{3-4}
			\multicolumn{2}{c|}{}                                      & 800$\times$19$\times$19   & 2$\times$2 max pool, stride 2                                                \\ \hline
			\multicolumn{2}{c|}{\begin{tabular}[c]{@{}c@{}}Dense Block\\ (3)\end{tabular}}                       & 1184$\times$19$\times$19  &                        \conv{8}                                               \\ \hline
			\multicolumn{2}{c|}{Transition w/o Pooling Layer (1)}      & 1184$\times$19$\times$19  & 1$\times$1 conv                                                              \\ \hline
			\multicolumn{2}{c|}{\begin{tabular}[c]{@{}c@{}}Dense Block\\ (4)\end{tabular}}                       & 1568$\times$19$\times$19  &                        \conv{8}                                               \\ \hline
			\multicolumn{2}{c|}{Transition w/o Pooling Layer (2)}      & 1568$\times$19$\times$19  & 1$\times$1 conv                                                              \\ \hline
			\multicolumn{2}{c|}{DSOD Prediction Layers}                           &    --       & Plain/Dense                                                           \\ \hline
		\end{tabular}
	}
\end{table*}

\section{Related Work}\label{related_work}
\vspace{.5em}
\noindent{\textbf{Object Detection.}}
Modern CNN-based object detectors can mainly be divided into two groups:
(i) proposal-based/two-stage methods; and (ii) proposal-free/one-stage methods.

Proposal-based family includes R-CNN~\cite{girshick2014rich}, Fast R-CNN~\cite{girshick2015fast}, Faster R-CNN~\cite{ren2015faster}, R-FCN~\cite{li2016r} and Mask RCNN~\cite{he2017mask}.
R-CNN uses selective search~\cite{uijlings2013selective} to first generate potential object regions in an image and then perform classification on the proposed regions. 
R-CNN requires high computational costs since each region is processed by the CNN network separately. Fast R-CNN improves the efficiency by sharing computation of backbone networks and Faster R-CNN uses neural networks (i.e., RPN) to generate the region proposals. R-FCN further improves speed and accuracy by removing fully-connected layers and adopting position-sensitive score maps for final detection.

Recently, in order to realize real-time object detection, the proposal-free methods like YOLO~\cite{redmon2016you} and SSD~\cite{liu2016ssd} have been proposed. YOLO uses a single feed-forward convolutional network to predict object classes and locations directly, which no longer requires a second per-region classification operation so that it is extremely fast. SSD further improves YOLO in several aspects, including (1) use small convolutional filters to predict categories and anchor offsets for bounding box locations; (2) use pyramid features for prediction at different feature scales; and (3) use default boxes and aspect ratios for adjusting varying object shapes. 
Some other proposal-free detectors also be proposed recently, e.g. RetinaNet~\cite{lin2017focal}, Scale-Transferrable~\cite{zhou2018scale}, Single-shot Refinement~\cite{zhang2018single}, RFB Net~\cite{liu2018receptive}, CornetNet~\cite{law2018cornernet}, ExtremeNet~\cite{zhou2019bottom}, etc.
Our proposed DSOD is built upon SSD framework and thus it inherits the speed and accuracy advantages of SSD, while produces more compact and flexible models.

\vspace{.5em}
\noindent{\textbf{Network Architectures for Detection.}}
Since there are significant efforts that have been devoted to design network architectures for image classification,
many diverse and powerful networks are emerged, such as AlexNet~\cite{krizhevsky2012imagenet}, VGGNet~\cite{simonyan2014very}, GoogLeNet~\cite{szegedy2015going}, ResNet~\cite{he2016deep}, DenseNet~\cite{huang2016densely}, etc. Meanwhile, several advanced regularization techniques~\cite{srivastava2014dropout,ioffe2015batch} also have been proposed to further enhance the model capabilities.
In practice, most of the detection methods \cite{girshick2014rich,girshick2015fast,ren2015faster,liu2016ssd} directly utilize these structures pre-trained on ImageNet as the backbone network for detection task.

Some other works try to design specific backbone network structures for object detection, but still require to pre-train on ImageNet classification dataset in advance.
Specifically, YOLO~\cite{redmon2016you} defines a network with 24 convolutional layers followed by 2 fully-connected layers. YOLO9000~\cite{redmon2016yolo9000} improves YOLO by proposing a new network named Darknet-19, which is a simplified version of VGGNet~\cite{simonyan2014very}. YOLOv3~\cite{redmon2018yolov3} further improve the performance through involving residual connection on Darknet-19 and other techniques.
Kim {\em et al.} \cite{kim2016pvanet} proposes PVANet for fast object detection, which consists of the simplified ``Inception'' block from GoogleNet.
Huang {\em et al.}~\cite{huang2016speed} investigated various combination of network structures and detection frameworks, and found that Faster R-CNN~\cite{ren2015faster} with Inception-ResNet-v2~\cite{szegedy2016inception} achieved very promising performance.
In this paper, we also consider designing a suitable backbone structure for generic object detection. 
However, the pre-training operation on ImageNet is no longer required by the proposed DSOD. 

\vspace{.5em}
\noindent{\textbf{Learning Deep Models from Scratch.}}
To the best of our knowledge, there are no previous works that train deep CNN-based object detectors from scratch.
Thus, our proposed approach has very appealing advantages over existing solutions.
We will elaborate and validate the method in the following sections.
%
In semantic segmentation, J{\'e}gou {\em {et al.}}~\cite{jegou2016one} demonstrated that a well-designed network structure can outperform state-of-the-art solutions without using the pre-trained models. It extends DenseNets to fully-convolutional networks by adding an upsampling path to recover the full input resolution. 

\section{DSOD}\label{model}
In this section, we first introduce the whole framework of our DSOD architecture, following by several important design principles. Then we describe the objective function and training settings in detail.

\vspace{-.8em}
\subsection{Network Architecture}

Similar to SSD \cite{liu2016ssd}, our proposed DSOD method is a multi-scale and proposal-free detection framework.
The network structure of DSOD can be divided into two parts: the backbone sub-network for feature extraction and the front-end sub-network for prediction over multi-resolution feature maps.
The backbone sub-network is a variant of the deeply supervised DenseNets \cite{huang2016densely} structure, which is composed of a {\em {stem block}}, four {\em {dense blocks}}, two {\em {transition layers}} and two {\em {transition w/o pooling layers}}.
The front-end subnetwork (or named {\em DSOD prediction layers}) fuses multi-scale prediction responses with an elaborated {\em {dense structure}}.
Fig.~\ref{dense_connection} illustrates the proposed DSOD prediction layers along with the plain structure used in SSD \cite{liu2016ssd}.
The full DSOD network architecture\footnote{The visualization of the complete network structure is available at: \url{http://ethereon.github.io/netscope/\#/gist/b17d01f3131e2a60f9057b5d3eb9e04d}.} is detailed in Tab.~\ref{base_network}.
Now we elaborate each component and the corresponding design principle in the following.

\subsection{Design Principles} \label{32}

\vspace{.8em}
\noindent{\textbf{Principle 1: Proposal-free.}} 
In order to reveal the potential influences in learning object detection from scratch, we investigated all the state-of-the-art CNN-based object detectors under the default settings.
As aforementioned, R-CNN and Fast R-CNN require external object proposal generators like selective search. Faster R-CNN and R-FCN require integrated region-proposal-network (RPN) to generate relatively fewer region proposals. YOLO and SSD are single-shot and proposal-free methods (one-stage), which handle object location and bounding box coordinates as a regression problem.
We observe that only proposal-free methods (one-stage detectors) can converge successfully without the pre-trained models if we follow the original settings without involving some significantly modifications (e.g., replacing RoI pooling with RoI align~\cite{he2017mask}, adopting Sync BN~\cite{peng2017megdet} or Group Norm~\cite{wu2018group} to mitigate small batch-size issue, etc.).
We conjecture this is due to the RoI pooling (Regions of Interest) in the other two categories of methods ---
RoI pooling uses quantization to generate features for each region proposals, which causes misalignments that hinders/reduces the gradients being smoothly back-propagated from region-level to convolutional feature maps.
The proposal-based methods work well with pre-trained network models because the parameter initialization is good for those layers before RoI pooling, while this is not true for training from scratch.

Hence, we arrive at the first principle: training detection network from scratch requires a proposal-free framework, even if there is no BN layer~\cite{ioffe2015batch} included in the network structures (In contrast, norm layer is critical for both Sync BN~\cite{peng2017megdet} and Group Norm~\cite{wu2018group} methods to train region-based/two-stage detectors from scratch).
In practice, we derive a multi-scale proposal-free framework from the SSD framework \cite{liu2016ssd}, as it could reach state-of-the-art accuracy while offering fast processing speed.

\vspace{.5em}
\noindent{\textbf{Principle 2: Deep Supervision.}}
Using deeply supervised structures to improve network performance has been demonstrated a effective practice in GoogLeNet~\cite{szegedy2015going}, DSN~\cite{lee2015deeply}, DeepID3~\cite{sun2015deepid3}, etc. Among these network structures,
the central idea is to provide integrated objective function as direct supervision to the earlier hidden layers, rather than only at the output one. 
These ``companion''  or ``auxiliary'' objective functions at multiple hidden layers can mitigate the ``vanishing'' gradients problem.
The proposal-free detection framework contains both classification and localization loss.
The explicit solution requires adding complex side-output layers to introduce ``companion'' objective at each hidden layer for the detection task, similar to \cite{xie2015hed}.
In this work, we empower \textit{deep supervision} with an elegant \& implicit solution called layer-wise dense connections, as introduced in DenseNets~\cite{huang2016densely}.
A block is called \textit{dense block} when all preceding layers in the block are connected to the current layer. 
Hence, earlier layers in DenseNet can receive additional supervision from the objective function through the skip connections.
Although only a single loss function is required on top of the network, all layers including the earlier layers still can share the supervised signals unencumbered.

In order to further verify the effectiveness of Deep Supervision mechanism, we propose a deep-scale supervised (DSS) module, which is similar to Hypernet~\cite{kong2016hypernet}, Inside-outside net~\cite{bell2016inside}, etc. As illustrated in Fig.~\ref{ds}, DSS concatenates three different scales of feature maps (low, middle and high levels) from different blocks into a single prediction module. For low-level (coarse resolution) features, we use a $4\times4$ max pooling, stride $=$ 2 to reduce the resolution, following by a $1\times1$ conv-layer for reducing the number of feature maps. We use the $2\times2$ max pooling for middle level feature maps and do not include max pooling for high-level layers. Then, we concatenate these diverse feature maps together for final prediction. Each prediction layer can be formulated as:
\begin{equation}
{\mathbb{P}_i} = {\phi _i}[{\textbf{P}_{1/4}}({x_L}),{\textbf{P}_{1/2}}({x_M}),{x_H}],
\end{equation}
where ${\mathbb{P}_i} (i=1,2,\cdots,5)$ denotes the $i$-th prediction layer outputs. $\textbf{P}_{1/k}$ denotes $k\times k$ max pooling. $x_L$, $x_M$ and $x_H$ denote feature maps from different layers.
We will verify the benefit of deep supervision in Section~\ref{validation}.

\vspace{.5em}
\noindent{\emph{Transition w/o Pooling Layer.}}
In order to increase the number of dense blocks without reducing the final feature map resolution, we introduce a new layer called {\emph {transition w/o pooling layer}.
In the original design of DenseNet, each transition layer contains a pooling operation to down-sample the feature resolution. The number of dense blocks is fixed (4 dense blocks in all DenseNet architectures) if one wants to maintain the same scale/size of outputs.
The only way to increase network depth is adding layers inside each block for the original DenseNet.
The {\emph{transition w/o pooling layer}} eliminates this restriction of the number of dense blocks in DSOD architecture. You can include any number of blocks in a network as you want, which can also be adopted by the standard DenseNet.

\begin{figure}[t]
	\vspace{-0.12in}
	\centering
	\includegraphics[width=0.23\textwidth]{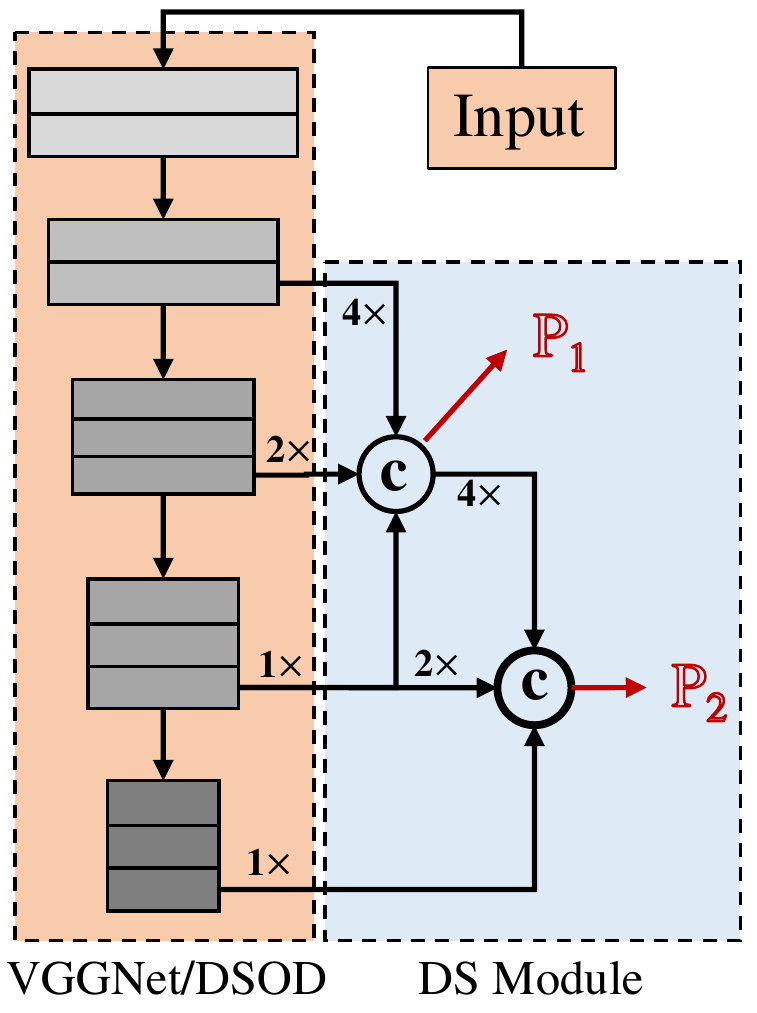}
	\vspace{-0.02in}
	\caption{Illustration of the deep-scale supervision (DSS) module. ``\textbf{4$\times$}, \textbf{2$\times$} and \textbf{1$\times$}'' denote that we reduce the resolution of feature maps to \textbf{$1/4$}, \textbf{$1/2$} and the original size, respectively. ``\textbf{c}'' denotes concatenation operation. ``\textbf{P$_1$} and \textbf{P$_2$}'' are the first (38$\times$38) and second scales (19$\times$19) of prediction modules in Fig.~\ref{dense_connection}. ``\textbf{P$_3$-P$_5$}'' also use three-scale feature maps for prediction, which are not presented in this Figure.}
	\label{ds}
	\vspace{-0.12in}
\end{figure}

\vspace{.5em}
\noindent{\textbf{Principle 3: Stem Block.}}
Motivated by Inception-v3~\cite{szegedy2016rethinking} and v4~\cite{szegedy2016inception}, we define stem block as a stack of three 3$\times$3 convolution layers followed by a 2$\times$2 max pooling layer. The first conv-layer works with stride = 2 and the other two are with stride = 1. We find that adding this simple stem structure can evidently improve the detection performance in our experiments.
We conjecture that, compared with the original design in DenseNet (7$\times$7 conv-layer, stride = 2 followed by a 3$\times$3 max pooling, stride = 2), the stem block can reduce the information loss from raw input images with small kernel size at the beginning of a network.
We will show that the reward of this stem block is significant for object detection performance in Section~\ref{validation}.

\vspace{.5em}
\noindent{\textbf{Principle 4: Dense Prediction Structure.}}
Fig.~\ref{dense_connection} illustrates the comparison of the plain structure (as in SSD) and our proposed dense structure in the front-end sub-network.
SSD designs prediction-layers as an asymmetric hourglass structure. For 300$\times$300 input size, SSD applies six scales of feature maps for predicting objects
.
The Scale-1 feature maps are from the middle layer of the backbone sub-network, which has the largest resolution (38$\times$38) in order to handle the small objects in an image.
The remaining five scales are on top of the backbone sub-network. Then, a plain transition layer with the {\em {bottleneck}} structure (a 1$\times$1 conv-layer for reducing the number of feature maps plus a 3$\times$3 conv-layer)~\cite{szegedy2016rethinking,he2016deep} is adopted between two contiguous scales of feature maps.

\vspace{.5em}
\noindent{\emph{Learning Half and Reusing Half.}} 
In plain structure, each later scale of prediction layer is directly transited from the adjacent previous scale layer, as shown in Fig.~\ref{dense_connection}, which is used in SSD framework.
In this work, we propose to use dense structure for prediction. Each prediction layer combines multi-scale information from two stages of layers.
For simplicity, we restrict that each scale outputs the same number of channels for the prediction feature maps as is in the plain structure.
In DSOD of each scale (except scale-1), half of the feature maps are learned from the previous scale layer with a series of conv-layers,
while the remaining half feature maps are directly down-sampled from the contiguous high-resolution feature maps.
The down-sampling block consists of a 2$\times$2, stride $=$ 2 max pooling layer followed by a 1$\times$1, stride = 1 conv-layer.
The pooling layer aims to match resolution to current size during concatenation.
The 1$\times$1 conv-layer is used to reduce the number of channels to 50\%.
The pooling layer is placed before the 1$\times$1 conv-layer for the consideration of reducing computing cost.
This down-sampling block actually brings each scale with the multi-resolution feature maps from all of its preceding scales,
which is essentially identical to the dense layer-wise connection introduced in DenseNets.
For each scale, we only learn half of new feature maps and reuse the remaining half of the previous ones.
This dense prediction structure can yield more accurate results with fewer parameters than the plain structure, as will be studied in Section~\ref{ablation}.

\begin{table}[]
	\centering
    \caption{\textbf{Effectiveness of various designs on VOC 2007 \texttt{test} set.} Please refer to Tab.~\ref{ablation_study} and Section~\ref{ablation} for more details.}
	\label{effects}
	\resizebox{1\linewidth}{!}{%
		\begin{tabular}{r|ccccccccc}
			
			& \multicolumn{7}{c}{DSOD300}                 \\ \hline
			transition w/o pooling? &    & \Checkmark    & \Checkmark  & \Checkmark  & \Checkmark  & \Checkmark  & \Checkmark  & \Checkmark & \Checkmark \\
			hi-comp factor $\theta$?         &  &      & \Checkmark  & \Checkmark  & \Checkmark  & \Checkmark  & \Checkmark  & \Checkmark & \Checkmark \\
			wide bottleneck?   &     &  &      & \Checkmark  & \Checkmark  & \Checkmark  & \Checkmark  & \Checkmark & \Checkmark \\
			wide 1st conv-layer? &    &   &      &      & \Checkmark  & \Checkmark  & \Checkmark  & \Checkmark & \Checkmark  \\
			big growth rate?               &    &      &      &      &      & \Checkmark  & \Checkmark  & \Checkmark & \Checkmark \\
			stem block?                           &     &      &      &      &      &      & \Checkmark  & \Checkmark & \Checkmark \\
			dense pred-layers?     & &       &      &      &      &      &      & \Checkmark & \Checkmark  \\
           DSS module?   & &       &      &      &      &      &      &  & \Checkmark  \\
			\hline
			VOC 2007 mAP & 59.9  & 61.6 & 64.5 & 68.6 & 69.7 & 74.5 & 77.3 & 77.7 & 79.1 \\
		\end{tabular}
	}
\end{table}

\renewcommand{\arraystretch}{1.03}
\setlength{\tabcolsep}{1.0em}
\begin{table*}[]
	\centering
    \caption{\textbf{Ablation study on PASCAL VOC 2007 \texttt{test} set.} \textbf{DS/A-B-$k$-$\theta$} describes our backbone network structure. \textbf{A} denotes the number of channels in the 1st conv-layer. \textbf{B} denotes the number of channels in each {\em{bottleneck}} layer (1$\times$1 convolution). \textbf{$k$} is the growth rate in dense blocks. \textbf{$\theta$} denotes the compression factor in transition layers. See Section~\ref{ablation} for more explanations.}
	\label{ablation_study}
	\resizebox{1.0\textwidth}{!}{%
		\begin{tabular}{c|c|c|c|c|c|c|c|c}
			\hline
			Method     & data  & pre-train & transition  w/o pool & stem & backbone & prediction layer & \# parameters &  (\%) mAP  \\ \hline
			DSOD300   & 07+12 &\ding{55}& \ding{55} &\ding{55}& DS/32-12-16-0.5 &   Plain  & 4.1M & 59.9 \\
			DSOD300   & 07+12 &\ding{55}&\Checkmark&\ding{55}& DS/32-12-16-0.5 &   Plain  & 4.2M & 61.6 \\
			DSOD300   & 07+12 &\ding{55}&\Checkmark&\ding{55}& DS/32-12-16-1 &   Plain  &   5.5M   & 64.5 \\
			DSOD300   & 07+12 &\ding{55}&\Checkmark&\ding{55}&  DS/32-64-16-1  &   Plain & 6.1M & 68.6 \\
			DSOD300   & 07+12 &\ding{55}&\Checkmark&\ding{55}&  DS/64-64-16-1  &   Plain  &   6.3M   & 69.7 \\
			DSOD300   & 07+12 &\ding{55}&\Checkmark&\ding{55}&  DS/64-192-48-1  &   Plain & 18.0M &  74.5\\
			\hline		
			DSOD300   & 07+12 &\ding{55}&\Checkmark&\Checkmark&  DS/64-12-16-1  &   Plain  &   5.2M   & 70.7 \\
			DSOD300   & 07+12 &\ding{55}&\Checkmark&\Checkmark&  DS/64-36-48-1  &   Plain  &   12.5M   & 76.0 \\
			DSOD300   & 07+12 &\ding{55}&\Checkmark&\Checkmark& DS/64-192-48-1  &   Plain  &  18.2M    & 77.3 \\ \hline
			DSOD300   & 07+12 &\ding{55}&\Checkmark&\Checkmark& DS/64-64-16-1  &   Dense &  5.9M  & 73.6 \\
			DSOD300   & 07+12 &\ding{55}&\Checkmark&\Checkmark& DS/64-192-48-1  &   Dense &  14.8M  & 77.7 \\
			DSOD300   & 07+12+COCO &\ding{55}&\Checkmark&\Checkmark& DS/64-192-48-1  &   Dense &  14.8M  & 81.7 \\
			\hline
		\end{tabular}
	}
\end{table*}

\vspace{-.6em}
\subsection{Training Objective}

Our whole training objective loss is derived from SSD~\cite{liu2016ssd} and Fast RCNN~\cite{girshick2015fast}, which is a weighted sum of the classification loss (cls) and the localization loss (reg):
\begin{equation}
L(p,{p^*},r,{g}) = \frac{1}{N}({L_{cls}}(p,{p^*}) + \alpha {p^*}{L_{reg}}(r,{g}))
\end{equation}
where $p$ denotes a discrete probability distribution that is computed by a softmax over the K+1 outputs. $p^*$ is the ground-truth class. $r$ is the bounding-box regression offsets and $g$ is the ground-truth bounding-box regression target. $\alpha$ is the coefficient to balance the two losses. Following Fast RCNN~\cite{girshick2015fast}, we also adopt the $L_1$ loss for bounding-box regression:

\begin{equation}
{L_{reg}}(r,{g}) = \sum\limits_{i \in \{ x,y,w,h\} } {smoot{h_{{L_1}}}} ({r_i} - g_i)
\end{equation}
Specially, we calculate the four coordinates following~\cite{liu2016ssd,girshick2015fast,ren2015faster}:
\begin{equation}
\begin{gathered}
  {r_x} = (x - {x_a})/{w_a},\;\;{r_y} = (y - {y_a})/{h_a}, \hfill \\
  {r_w} = \log (w/{w_a}),\;\;{r_h} = \log (h/{h_a}), \hfill \\
  g_x = ({x^*} - {x_a})/{w_a},\;\;g_y = ({y^*} - {y_a})/{h_a}, \hfill \\
  g_w = \log ({w^*}/{w_a}),\;\;g_h = \log ({h^*}/{h_a}), \hfill \\ 
\end{gathered} 
\end{equation}
where $x$, $y$, $w$, and $h$ denote the box’s center coordinates and its width and height. $x$, $x_a$ and $x^*$ denote predicted box, default box and ground-truth box, respectively.

\vspace{-.8em}
\subsection{Other Settings}
We implement our detectors based on the caffe platform~\cite{jia2014caffe}.
All our models are trained from scratch with SGD solver on NVidia TitanX GPU.
Since each scale of DSOD feature maps is concatenated from multi-resolution features, we adopt L2 normalization technique~\cite{liu2015parsenet} to scale the feature norm to 20 on all outputs.
Note that SSD only applies this normalization to scale-1. Most of our training strategies follow SSD, including data augmentation, scale and aspect ratios for default boxes, etc.,
while we have our own learning rate scheduling and mini-batch size settings. Details will be given in the experimental section.

\renewcommand{\arraystretch}{1.03}
\setlength{\tabcolsep}{1.0em}
\begin{table*}[]
	\centering
    \caption{\textbf{PASCAL VOC 2007 \texttt{test} detection results.} SSD300* is updated version by the authors after the paper publication. SSD300S$^\dagger$ indicates training SSD300* from scratch with ResNet-101 or VGGNet, which serves as our baseline. Note that the speed of Faster R-CNN with  ResNet-101 (2.4 \textit{fps}) is tested on K40, while others are tested on Titan X. The result of SSD300S with ResNet-101 (63.8\% mAP, without the pre-trained model) is produced with the default setting of SSD, which may not be optimal.} 
	\label{VOC2007}
	\resizebox{1\textwidth}{!}{%
		\begin{tabular}{c|c|c|c|c|c|c|c|c}
			\hline
			Method     & data  & pre-train &    backbone   & prediction layer  & speed (\textit{fps}) & \# parameters & input size & (\%) mAP  \\ 
  			\hline
			Faster RCNN~\cite{ren2015faster}  & 07+12 &\Checkmark&     VGGNet    &     -      &   7    & 134.7M & $\sim600\times1000$&73.2 \\
			Faster RCNN~\cite{ren2015faster}  & 07+12 &\Checkmark&   ResNet-101    &     -      &  2.4$^*$     &   -   &$\sim600\times1000$& 76.4 \\
			R-FCN~\cite{li2016r}     & 07+12 &\Checkmark&   ResNet-50    &     -      &    11   &31.9M& $\sim600\times1000$& 77.4 \\
			R-FCN~\cite{li2016r}     & 07+12 &\Checkmark&   ResNet-101    &     -      &  9     &50.9M& $\sim600\times1000$& 79.5 \\
			R-FCN{\footnotesize{multi-sc}}~\cite{li2016r}    & 07+12 &\Checkmark&ResNet-101& - & 9   &50.9M&$\sim600\times1000$&  80.5 \\ \hline
			YOLOv2~\cite{redmon2016yolo9000}   & 07+12 &\Checkmark&  Darknet-19       &   -    &     81  &-&$352\times352$&  73.7 \\
			SSD300~\cite{liu2016ssd}   & 07+12 &\Checkmark&  VGGNet       &   Plain    &     46  &26.3M&$300\times300$&  75.8 \\
			SSD300*~\cite{liu2016ssd}   & 07+12 &\Checkmark&   VGGNet       &   Plain    &    46   &26.3M& $300\times300$& 77.2 \\
			\hline
			Faster RCNN  & 07+12 &\ding{55}&  VGGNet/ResNet-101/DenseNet  &   \multicolumn{5}{c}{Failed}    \\
			R-FCN     & 07+12 &\ding{55}&  VGGNet/ResNet-101/DenseNet  &   \multicolumn{5}{c}{Failed}    \\ \hline
			SSD300S$^\dagger$    & 07+12 &\ding{55}&   ResNet-101    &   Plain    &   12.1    &   52.8M   & $300\times300$& 63.8$^*$ \\
			SSD300S$^\dagger$     & 07+12 &\ding{55}&    VGGNet       &   Plain    &   46    &  26.3M    &$300\times300$&  69.6 \\
			SSD300S$^\dagger$     & 07+12 &\ding{55}&    VGGNet       &   Dense    &   37    &  26.0M    &$300\times300$&  70.4 \\ \hline
			DSOD300   & 07+12 &\ding{55}&  DS/64-192-48-1  &   Plain    &  20.6     &  18.2M    &$300\times300$&  77.3 \\
			DSOD300   & 07+12 &\ding{55}&  DS/64-192-48-1  &   Dense   &   17.4    &  14.8M  &$300\times300$&  77.7 \\
			DSOD300   & 07+12+COCO &\ding{55}&  DS/64-192-48-1  &   Dense &   17.4  &  14.8M &$300\times300$& 81.7 \\ \hline

		\end{tabular}
	}
\end{table*}

\section{Experiments}\label{experiments}
Our experiments are conducted on the widely used PASCAL VOC 2007, 2012 and MS COCO datasets that have 20, 20, 80 object categories respectively.
We adopt the standard mean Average Precision (mAP) to measure the object detection performance.

\vspace{-.8em}
\subsection{Ablation Study on PASCAL VOC2007} \label{ablation}
We first investigate each component and design principle of our DSOD framework. The results are mainly summarized in Tab.~\ref{effects} and Tab.~\ref{ablation_study}.
We design several controlled experiments on PASCAL VOC 2007 with our DSOD300 (with 300$\times$300 inputs) for this ablation study.
A consistent setting is imposed on all the experiments, unless when some components or structures are examined.
In this study, we train the models with the combined training set from VOC 2007 \texttt{trainval} and 2012 \texttt{trainval} (``07+12''), and test on the VOC 2007 \texttt{test} set.

\subsubsection{Configurations in Dense Blocks}
In this section, we first investigate the impact of different configurations in dense blocks of the backbone sub-network.

\vspace{.5em}
\noindent{\textbf{Compression Factor in Transition Layers.}}
We compare two compression factor values ($\theta$ = 0.5, 1) in the transition layers of DenseNets.
Results are shown in Tab.~\ref{ablation_study} (rows 2 and 3).
Compression factor $\theta$ = 1 means that there is no feature map reduction in the transition layer, while $\theta$ = 0.5 means half of the feature maps are reduced.
We can observe that $\theta$ = 1 obtains 2.9\% higher mAP than $\theta$ = 0.5.

\vspace{.5em}
\noindent{\textbf{\# Channels in bottleneck layers.}}
As shown in Tab.~\ref{ablation_study} (rows 3 and 4),
we observe that wider bottleneck layers (with more channels of response maps) improve the performance greatly (4.1\% mAP).

\vspace{.5em}
\noindent{\textbf{\# Channels in the 1st conv-layer}}
We observe that a large number of channels in the first conv-layers is beneficial, which brings 1.1\% mAP improvement (in Tab.~\ref{ablation_study} rows 4 and 5).

\vspace{.5em}
\noindent{\textbf{Growth rate.}}
A large growth rate $k$ is found to be much better.
We observe 4.8\% mAP improvement in Tab.~\ref{ablation_study} (rows 5 and 6) when increase $k$ from 16 to 48 with 4$k$ bottleneck channels.

\renewcommand{\arraystretch}{1.03}
\setlength{\tabcolsep}{1.0em}
\begin{table*}[]\small
	\centering
    \caption{\textbf{PASCAL VOC 2012 \texttt{test} detection results.}
	\textbf{07+12}: 07 \texttt{trainval} + 12 \texttt{trainval}, \textbf{07+12+S}: 07+12 plus segmentation labels, \textbf{07++12}: 07 \texttt{trainval} + 07 \texttt{test} + 12 \texttt{trainval}. Anonymous result links are DSOD300 (07+12) : \url{http://host.robots.ox.ac.uk:8080/anonymous/PIOBKI.html}; DSOD300 (07+12+COCO): \url{http://host.robots.ox.ac.uk:8080/anonymous/I0UUHO.html}.} 
	\label{voc12}
	\setlength{\tabcolsep}{1.82pt}
	\resizebox{1\textwidth}{!}{%
		\begin{tabular}{l|c|c|c|c|cccccccccccccccccccc}
			\hline
			Method &  data  &  backbone &  pre-train &  mAP &  aero &  bike &  bird &  boat &  bottle &  bus &  car &  cat &  chair &  cow &  table &  dog &  horse &  mbike &  person &  plant &  sheep &  sofa &  train &  tv \\
			\hline
			HyperNet~\cite{kong2016hypernet} &  07++12 &  VGGNet & \Checkmark& 71.4 &  84.2&78.5&73.6& 55.6& 53.7& 78.7& 79.8& 87.7& 49.6& 74.9& 52.1& 86.0& 81.7& 83.3& 81.8& 48.6& 73.5& 59.4& 79.9& 65.7 \\
			ION~\cite{bell2016inside} &  07+12+S &  VGGNet & \Checkmark& 76.4 & 87.5 & 84.7 & 76.8 & 63.8 & 58.3 & 82.6 & 79.0 & 90.9 & 57.8 & 82.0 & 64.7 & 88.9 & 86.5 & 84.7 & 82.3 & 51.4 & 78.2 & 69.2 & 85.2 & 73.5 \\
			Faster RCNN~\cite{ren2015faster} &  07++12 &  ResNet-101 & \Checkmark& 73.8 & 86.5 & 81.6 & 77.2 & 58.0 & 51.0 & 78.6 & 76.6 & 93.2 & 48.6 & 80.4 & 59.0 & 92.1 & 85.3 & 84.8 & 80.7 & 48.1 & 77.3 & 66.5 & 84.7 & 65.6 \\
			R-FCN{\footnotesize{multi-sc}}~\cite{li2016r} &  07++12&  ResNet-101 & \Checkmark& 77.6 & 86.9 & 83.4 & 81.5& 63.8& 62.4 & 81.6 & 81.1 & 93.1 & 58.0 & 83.8 & 60.8& 92.7 & 86.0 & 84.6 & 84.4 & 59.0 & 80.8 & 68.6& 86.1 & 72.9 \\ \hline
			YOLOv2~\cite{redmon2016yolo9000} &  07++12 &  Darknet-19 & \Checkmark& 73.4 & 86.3 & 82.0 & 74.8 & 59.2 & 51.8 & 79.8 & 76.5 & 90.6 & 52.1 & 78.2 & 58.5 & 89.3 & 82.5 & 83.4 & 81.3 & 49.1 & 77.2 & 62.4 & 83.8 & 68.7 \\
			SSD300*~\cite{liu2016ssd} &  07++12&
			VGGNet & \Checkmark& 75.8 & 88.1 & 82.9 & 74.4 & 61.9 & 47.6 & 82.7 & 78.8 & 91.5 & 58.1 & 80.0 & 64.1 & 89.4 & 85.7 & 85.5 & 82.6 & 50.2 & 79.8 & 73.6 & 86.6 & 72.1\\	
			\hline	
			DSOD300 &  07++12&
			DS/64-192-48-1 & \ding{55} & 76.3 & 89.4  & 85.3 & 72.9 & 62.7 & 49.5 & 83.6 & 80.6 & 92.1 & 60.8 & 77.9 & 65.6 & 88.9 & 85.5 & 86.8 & 84.6 & 51.1 & 77.7 & 72.3 & 86.0 & 72.2 \\
			DSOD300 &  07++12+COCO&
			DS/64-192-48-1 & \ding{55} & 79.3 &  90.5 & 87.4 & 77.5 & 67.4 & 57.7 & 84.7 & 83.6 & 92.6 & 64.8 & 81.3 & 66.4 & 90.1 & 87.8 & 88.1 & 87.3 & 57.9 & 80.3 & 75.6 & 88.1 & 76.7 \\
			\hline	
		\end{tabular}
	}
\end{table*}

\subsubsection{Effectiveness of Design Principles} \label{validation}
In this section, we justify the effectiveness of each design principle elaborated earlier.

\vspace{.5em}
\noindent{\textbf{Proposal-free Framework.}}
We tried to learn object detectors from scratch using the proposal-based framework including Faster R-CNN and R-FCN with the default settings.
However, the training process failed to converge for all the network structures we attempted (VGGNet, ResNet, DenseNet).
We then tried to train with the proposal-free framework SSD.
The training converged successfully but still gave relatively worse results (69.6\% for VGGNet backbone) compared with the case fine-tuning from pre-trained model (75.8\%), as shown in Tab.~\ref{VOC2007}.
These experiments validate our principle to choose a proposal-free framework.

\begin{table}[]
	\centering
    \caption{\textbf{Effectiveness of various designs on VOC 2007 \texttt{test} set.} \textbf{DP} denotes dense prediction. \textbf{DSS w/o BN} denotes deep-scale supervision module without BN~\cite{ioffe2015batch}. Please refer to Section~\ref{ablation} for more details.}
	\label{effects}
	\resizebox{0.78\linewidth}{!}{%
		\begin{tabular}{l|c|c}
        \hline
Method & pre-train   & (\%) mAP \\ \hline \hline
SSD~\cite{liu2016ssd}  & \Checkmark  & 77.2 \\ \hline
SSD~\cite{liu2016ssd}  & \ding{55}  & 69.6 \\
SSD~\cite{liu2016ssd} (\textbf{+DP}) & \ding{55}  & 70.4 \\
SSD~\cite{liu2016ssd} (\textbf{+DP+DSS w/o BN}) & \ding{55}  & 74.2\\
SSD~\cite{liu2016ssd} (\textbf{+DP+DSS w/ BN}) & \ding{55}  & \textbf{77.4}\\
\hline \hline
DSOD  & \ding{55}  & 77.7 \\
DSOD (v2) (\textbf{+DSS w/ BN}) & \ding{55}  & \textbf{79.1}\\
\hline
\end{tabular}
	}
	\vspace{-0.1in}
\end{table}

\begin{figure*}[]
	\centering
	\includegraphics[width=0.8\textwidth]{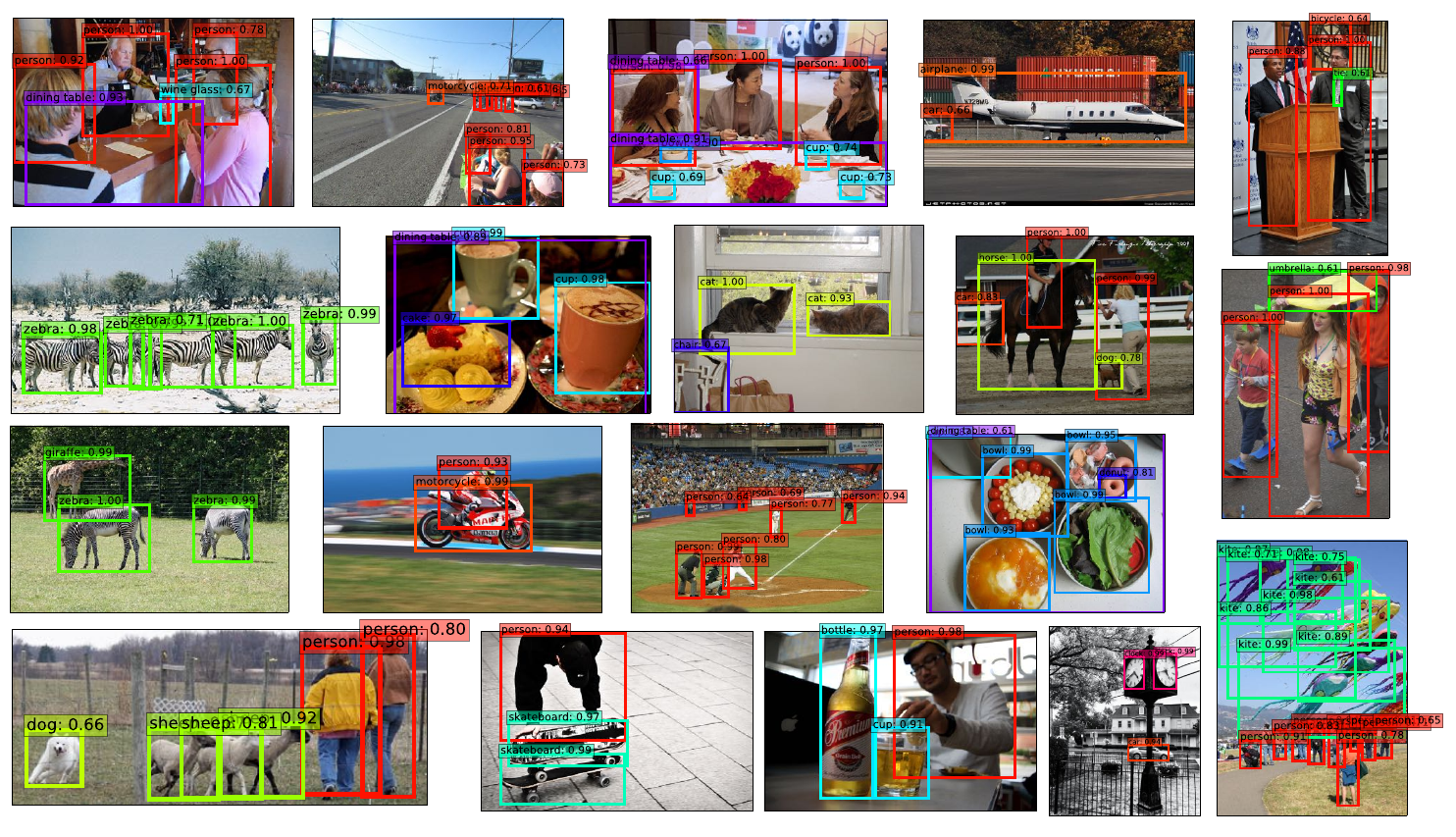}
	\vspace{-0.16in}
	\caption{Examples of object detection results on the MS COCO {\em {test-dev}} set using DSOD300. The training data is COCO {\em {trainval}} without the ImageNet pre-trained models (29.3\% mAP@[0.5:0.95] on the {\em {test-dev}} set). Each output box is associated with a category label and a softmax score in [0, 1]. A score threshold of 0.6 is used for displaying. For each image, one color corresponds to one object category in that image. The running time per image is 57.5ms on one Titan X GPU or 590ms on Intel (R) Core (TM) i7-5960X CPU @ 3.00GHz.}
	\label{examples}
\end{figure*}

\renewcommand{\arraystretch}{1.03}
\setlength{\tabcolsep}{1.0em}
\begin{table*}[t]\small
	\centering
    \caption{\textbf{PASCAL VOC 2012 Competition \texttt{Comp3} results.} The training data is PASCAL VOC 2012 \texttt{trainval} set without pre-trained models. Anonymous result link of DSOD v2 is \url{http://host.robots.ox.ac.uk:8080/anonymous/TOAZCG.html}.
	} 
	\label{leaderboard}
	\setlength{\tabcolsep}{1.82pt}
	\resizebox{0.93\textwidth}{!}{%
		\begin{tabular}{l|c|c|c|c|c|c|c|c|c|c|c|c|c|c|c|c|c|c|c|c|c}
			\hline
			Method &  mAP &  aero &  bike &  bird &  boat &  bottle &  bus &  car &  cat &  chair &  cow &  table &  dog &  horse &  mbike &  person &  plant &  sheep &  sofa &  train &  tv \\
			\hline
	\textbf{DSOD (v2)} & \textbf{72.9}	& \textbf{86.8} & \textbf{82.5} &	\textbf{69.0} &	\textbf{57.4} &	\textbf{47.1} &	\textbf{81.2} &	\textbf{77.8} &	\textbf{88.7} &	\textbf{54.8} &	\textbf{75.5} &	{60.4} &	\textbf{85.2} &	\textbf{82.0} &	\textbf{85.4} &	\textbf{82.4} &	\textbf{45.0} &	\textbf{75.3} &	\textbf{68.2} &	\textbf{84.3} & \textbf{69.2} \\ 
			{DSOD} & {70.8}	& {86.4} & {80.2} &	{65.5} &	{55.7} &	{42.4} &	{80.3} &	{75.3} &	{86.6} &	{51.1} &	{72.3} &	\textbf{60.5} &	{83.9} &	{80.5} &	{83.6} &	{80.4} &	{42.7} &	{72.4} &	{67.3} &	{83.1} & {66.2} \\ \hline
			SSD~\cite{liu2016ssd}  & 64.0 & 78.9  & 72.3 & 61.8 &42.8  &27.9  & 73.1 & 69.4 & 84.9 &42.5  & 68.4 & 52.2 & 80.9 & 76.5 & 77.2 & 68.2 & 31.6 & 67.0 & 66.6 & 77.3 & 60.9 \\
			THU\_ML\_class  & 62.4& 78.0  & 71.0 & 64.5 &47.4  &45.3  & 70.1 & 70.6 & 82.0 &37.9  & 65.4 & 44.2 & 77.4 & 69.6 & 74.4 & 75.5 & 37.9 & 62.0 & 45.5 & 73.8 & 56.3 \\
			YOLOv2~\cite{redmon2016yolo9000}  & 48.8 & 69.5  & 61.6 &37.6  &28.2  & 18.8 & 63.2 & 53.2 & 65.6 & 27.5 & 44.4 & 35.9 & 61.4 & 57.9 & 66.9 & 63.8 & 16.8 & 52.8 & 39.5 & 65.4 & 46.2 \\
			DENSE\_BOX  & 45.9 & 64.7  & 64.1 & 28.8  & 26.7  & 30.7 & 60.6 & 54.9 & 47.4 & 29.3 & 41.8 & 34.6 & 42.6 & 59.3 & 64.2 & 62.5 & 24.3 & 53.7 & 27.1 & 50.9 & 50.7 \\
			NoC  & 42.2 & 62.8  & 60.4 & 26.7  & 22.3  & 25.7 & 56.9 & 55.2 & 52.1 & 21.5 & 38.3 & 34.2 & 43.9 & 51.2 & 58.8 & 40.7 & 20.4 & 42.0 & 37.4 & 52.6 & 41.6 \\
			\hline	
		\end{tabular}
	}
	\vspace{.6ex}
\end{table*}

\begin{table*}[t]
	\centering
    \caption{\textbf{MS COCO \texttt{test-dev 2015} detection results.}}
	\label{COCO}
	\resizebox{1\textwidth}{!}{%
		\begin{tabular}{l|c|c|c|ccc|ccc|ccc|ccc}
			\hline
			\multirow{2}{*}{Method}          & \multirow{2}{*}{data} & \multirow{2}{*}{network} & \multirow{2}{*}{pre-train} & \multicolumn{3}{c|}{Avg. Precision, IoU:} & \multicolumn{3}{c|}{Avg. Precision, Area:} & \multicolumn{3}{c|}{Avg. Recall, \#Dets:} & \multicolumn{3}{c}{Avg. Recall, Area:} \\
			&                       &              &            & 0.5:0.95        & 0.5        & 0.75       & S            & M            & L            & 1            & 10           & 100         & S           & M           & L           \\ \hline
			Faster RCNN~\cite{ren2015faster} & trainval              & VGGNet          &    \Checkmark     & 21.9            & 42.7       & -          & -            & -            & -            & -            & -            & -           & -           & -           & -           \\
			ION~\cite{bell2016inside}  & train                 & VGGNet    &    \Checkmark   & 23.6            & 43.2       & 23.6       & 6.4          & 24.1         & 38.3         & 23.2         & 32.7         & 33.5        & 10.1        & 37.7        & 53.6        \\
			R-FCN~\cite{li2016r}        & trainval              & ResNet-101 & \Checkmark & 29.2            & 51.5       & -          & 10.3         & 32.4         & 43.3         & -            & -            & -           & -           & -           & -           \\
			R-FCN{\footnotesize{multi-sc}}~\cite{li2016r}        & trainval              & ResNet-101 & \Checkmark & 29.9            & 51.9       & -          & 10.8         & 32.8         & 45.0         & -            & -            & -           & -           & -           & -           \\ \hline	
			YOLOv2~\cite{redmon2016yolo9000}        & trainval35k           & Darknet-19   & \Checkmark      & 21.6            & 44.0       & 19.2       & 5.0          & 22.4         & 35.5         & 20.7         & 31.6         & 33.3        & 9.8        & 36.5        & 54.4        \\ 		
			SSD300*~\cite{liu2016ssd}        & trainval35k           & VGGNet   & \Checkmark      & 25.1            & 43.1       & 25.8       & 6.6          & 25.9         & 41.4         & 23.7         & 35.1         & 37.2        & 11.2        & 40.4        & 58.4        \\
			DSOD300                           & trainval         & DS/64-192-48-1 & \ding{55} & 29.3            & 47.3       & 30.6       & 9.4          & 31.5         & 47.0         & 27.3         & 40.7         & 43.0        & 16.7        & 47.1        & 65.0        \\ \hline
		\end{tabular}
	}
		\vspace{-0.15in}
\end{table*}

\vspace{.5em}
\noindent{\textbf{Deep Supervision.}}
We then tried to learn object detectors from scratch with the principle of deep supervision.
Our DSOD300 achieves 77.7\% mAP, which is much better than the SSD300S that
is trained from scratch using VGG16 (69.6\%) without deep supervision.
Since VGGNet is a plain network, we design a deep-scale supervision (DSS) module to further validate the effectiveness of deep supervision. The structure of our DSS is shown in Fig.~\ref{ds}, we can observe that DSS structure concatenates three different scales of feature maps (low, middle and high levels) into a single prediction module. The performance comparisons are shown in Tab.~\ref{effects}, our proposed module significantly improves the accuracy of SSD from 70.4\% to 77.4\%, even better than the ImageNet pre-trained case (77.2\%). Adopting DSS module in DSOD can obtain consistent improvement (79.1\%).
 

\vspace{.5em}
\noindent{\textbf{Transition w/o Pooling Layer.}}
We compare the case without this designed layer (only 3 dense blocks) and the case with the designed layer (4 dense blocks in our design).
The backbone network is DS/32-12-16-0.5. Results are shown in Tab.~\ref{ablation_study}. The network structure with the {\em Transition w/o pooling layer} leads deeper network structure and brings 1.7\% detection performance gain,
which validates the effectiveness of this layer.

\vspace{.5em}
\noindent{\textbf{Stem Block.}}
As shown in Tab.~\ref{ablation_study} (rows 6 and 9), the stem block notably improves the performance from 74.5\% to 77.3\%. 
This validates our conjecture that using stem block can protect information loss from the raw input images.

\vspace{.5em}
\noindent{\textbf{Dense Prediction Structure.}}
We analyze the dense prediction structure from three aspects: speed, accuracy and parameters.
As shown in Tab.~\ref{VOC2007}, DSOD with dense front-end structure runs slightly lower than the plain structure (17.4 \textit{fps} vs. 20.6 \textit{fps}) on a Titan X GPU, due to the overhead from additional down-sampling blocks.
However, the dense structure improves mAP from 77.3\% to 77.7\%,
meanwhile, it reduces the parameters from 18.2M to 14.8M.
Tab.~\ref{ablation_study} gives more details (rows 9 and 10).
We also tried to replace the prediction layers in SSD with the proposed dense prediction layers.
The accuracy on VOC 2007 \texttt{test} set can be improved from 75.8\% (original SSD) to 76.1\% (with pre-trained models), and 69.6\% to 70.4\% (w/o pre-trained models), when using the VGG-16 model as backbone.
This verifies the effectiveness of the dense prediction layer.

\vspace{.5em}
\noindent{\textbf{What happened if pre-training on ImageNet?}}
It is interesting to see the performance of DSOD with backbone network pre-trained on ImageNet.
We trained one lite backbone network DS/64-12-16-1 on ImageNet, which obtains 66.8\% top-1 accuracy and 87.8\% top-5 accuracy on the validation-set (slightly worse than VGG-16).
After fine-tuning the whole detection framework on ``07+12" \texttt{trainval} set, we achieve 70.3\% mAP on the VOC 2007 \texttt{test} set.
The comparison of corresponding training-from-scratch solution achieves 70.7\% accuracy, which is even slightly better.
We will further investigate this point more thoroughly in the future work.

\subsubsection{Runtime Analysis}\label{running_time}
The comprehensive inference speed comparisons are shown in the 6\emph{th} column of Tab.~\ref{VOC2007}.
With 300$\times$300 input, our DSOD can process an image in 48.6ms (20.6 \textit{fps}) on a single Titan X GPU with the plain prediction structure,
and 57.5ms (17.4 \textit{fps}) with the dense prediction structure.
As a comparison, R-FCN runs at 90ms (11 \textit{fps}) for ResNet-50 and 110ms (9 \textit{fps}) for ResNet-101.
The SSD300$^*$ runs at 82.6ms (12.1 \textit{fps}) for ResNet-101 and 21.7ms (46 \textit{fps}) for VGGNet.
In addition, our model uses about only 1/2 parameters to SSD300 with VGGNet, 1/4 to SSD300 with ResNet-101, 1/4 to R-FCN with ResNet-101 and 1/10 to Faster R-CNN with VGGNet. {A lite-version of DSOD (10.4M parameters, w/o any speed optimization) can run 25.8 \textit{fps} with only 1\% mAP drops.}

\vspace{-.8em}
\subsection{Results on PASCAL VOC2007}
Our models are trained based on the union of VOC 2007 \texttt{trainval} and VOC 2012 \texttt{trainval} (``07+12'') following~\cite{liu2016ssd}. We use a batch size of 128 cross 8 GPUs during training.
Note that this batch-size is beyond the capacity of GPU memories (even for an 8 GPU server,  each with 12GB memory).
We use a trick to overcome the GPU memory constraints by accumulating gradients over two training iterations, which has been implemented on Caffe platform~\cite{jia2014caffe}.
The initial learning rate is set to 0.1, and then divided by 10 after every 20k iterations.
The training finished when reaching 100k iterations. Following~\cite{liu2016ssd}, we use a weight decay of 0.0005 and a momentum of 0.9.
All conv-layers are initialized with the ``xavier'' method~\cite{glorot2010understanding}.

Tab.~\ref{VOC2007} shows our results on VOC2007 \texttt{test} set. SSD300$^*$ is the updated SSD results which use new data augmentation technique. Our DSOD300 with plain structure achieves 77.3\%, which is slightly better than SSD300$^*$ (77.2\%).
DSOD300 with dense prediction structure further improves the result to 77.7\%. 

\vspace{-.8em}
\subsection{Results on PASCAL VOC2012}
For VOC 2012 dataset, we use VOC 2012 {\texttt {trainval}} and VOC 2007 {\texttt {trainval}} + {\texttt{test}} for training, and test on VOC 2012 {\texttt{test}} set. The initial learning rate is set to 0.1 for the first 30k iterations, then divided by 10 after every 20k iterations.
The total training iterations are 110k. Other settings are the same as those used in our VOC 2007 experiments.
Our results of DSOD300 are shown in Tab.~\ref{voc12}. DSOD300 achieves 76.3\% mAP, which is consistently better than baseline SSD300$^*$ (75.8\%). 

\vspace{-.8em}
\subsection{Results on PASCAL VOC2012 Comp3}
VOC2012 Comp3 is the sub-challenge of PASCAL VOC 2012 which compares object detectors that are trained only with PASCAL VOC 2012 data (11,540 images in \texttt{trainval} set for training and 10,991 in \texttt{test} set for testing).

Our results are shown in Tab.~\ref{leaderboard}, DSOD achieves 70.8\% mAP on PASCAL VOC 2012 \texttt{test} set, which outperforms the baseline method SSD with a large margin (6.8\% mAP). DSOD v2 further improves the performance from 70.8\% to 72.9\% mAP.

 \begin{figure}[t]
 	\centering
 	\includegraphics[width=0.2\textwidth]{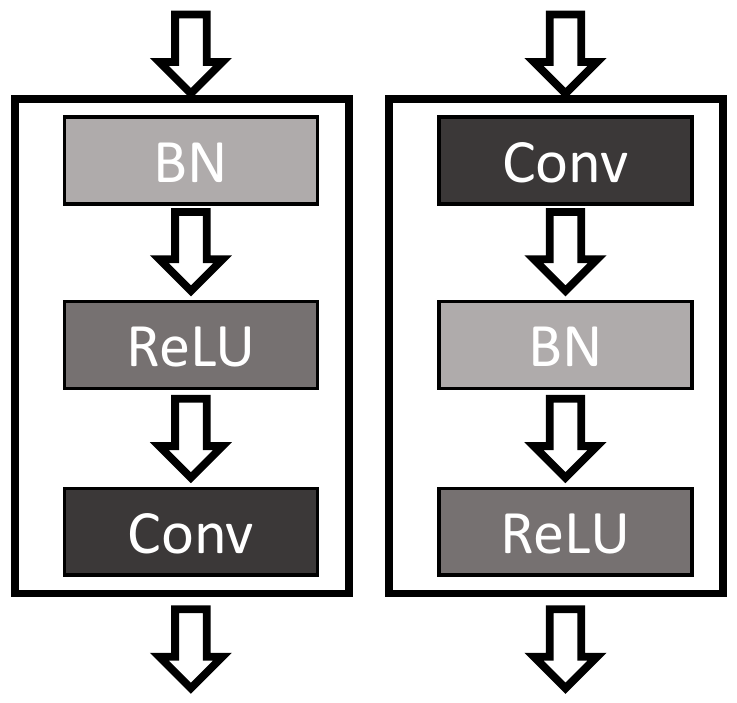}
 	\vspace{-0.12in}
 	\caption{Left is the pre-activation of BN-ReLU-Conv in DSOD, right is the post-activation of Conv-BN-ReLU in DSOD v2.}
 	\label{post_act}
 	\vspace{-0.18in}
 \end{figure}

\begin{figure*}[t]
	\centering
	\includegraphics[width=0.95\textwidth]{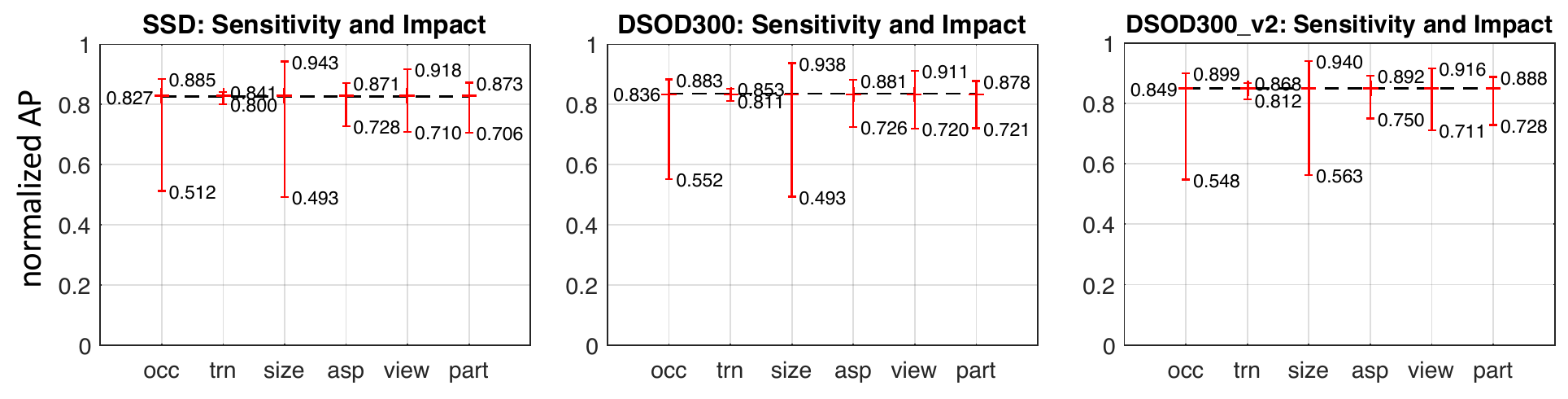}
	\vspace{-0.12in}
	\caption{Sensitivity of our detection results. Each plot shows the mean (over classes) normalized AP for the highest and lowest performing subsets within six different object characteristics (occlusion, truncation, bounding-box area, aspect ratio, viewpoint, part visibility). We show plots for our baseline method (SSD) and  our method (DSOD) with and without DSS module. We can observe that DSOD and DSOD v2 consistently improve the performance compared with baseline SSD.}
	\label{sensitivity}
	\vspace{-0.08in}
\end{figure*}

\begin{table*}[t]
	\centering
    \caption{\textbf{Comparisons of DSOD and DSOD (v2) on PASCAL VOC and MS COCO 2015 \texttt{test-dev} set.}}
    \vspace{-0.03in}
	\label{comparisons}
	\resizebox{0.8\textwidth}{!}{%
		\begin{tabular}{l|c|c|c|ccc}
			\hline
			\multirow{2}{*}{Method}          & \multirow{2}{*}{\textbf{VOC 07 (\% mAP)}} & \multirow{2}{*}{\textbf{VOC 12 (\% mAP)}} & \multirow{2}{*}{\textbf{VOC 12 Comp3 (\% mAP)}} & \multicolumn{3}{c}{\textbf{COCO (Avg. Precision, IoU:)}}  \\
			&                       &              &            & \textbf{0.5:0.95}        & \textbf{0.5}        & \textbf{0.75}          \\ \hline
			\textbf{DSOD300}   &   77.7   & 76.3  & 70.8 & 29.3            & 47.3       & 30.6       \\ \hline
           	\textbf{DSOD300* (v2)}   & \textbf{79.1}  & \textbf{77.2} & \textbf{72.9} & \textbf{30.4}  & \textbf{49.0}   & \textbf{31.8}  \\ \hline
		\end{tabular}
	}
\end{table*}

\begin{table*}[t]
	\centering
    \caption{\textbf{Comparisons of state-of-the-art two-stage detectors on MS COCO 2015 \texttt{test-dev} set. For fair comparisons, we resize the short side of inputs to 300 for all two-stage detectors. ``500'' indicates the max size of the inputs.}}
	\label{one_two}
    \vspace{-0.03in}
	\resizebox{0.8\textwidth}{!}{%
		\begin{tabular}{l|c|c|c|ccc}
			\hline
			\multirow{2}{*}{Method}          & \multirow{2}{*}{\textbf{network}} & \multirow{2}{*}{\textbf{pre-train}} & \multirow{2}{*}{\textbf{\# param}} & \multicolumn{3}{c}{\textbf{COCO (Avg. Precision, IoU:)}}  \\
			&                       &              &            & \textbf{0.5:0.95}        & \textbf{0.5}        & \textbf{0.75}          \\ \hline
           \multicolumn{7}{l}{\textbf{One-Stage Detectors:}}        \\ \hline
           \textbf{SSD300}~\cite{liu2016ssd}   &  VGGNet  & \Checkmark & 34.3M & 23.2      &  41.2    &  23.4      \\ 
			\textbf{SSD300*}~\cite{liu2016ssd}   &  VGGNet  & \Checkmark & 34.3M & 25.1           & 43.1       & 25.8       \\ 
			\textbf{DSOD300}   &   DSOD   & \ding{55}  & 21.8M & 29.3            & 47.3       & 30.6       \\ 
           	\textbf{DSOD300 (v2)}   &  DSOD + \textbf{DSS} & \ding{55} & 37.3M & \textbf{30.4}  & 49.0   & \textbf{31.8}  \\ \hline
         \multicolumn{7}{l}{\textbf{Two-Stage Detectors:}}        \\ \hline
            \textbf{FPN300/500}~\cite{lin2016feature}   &  ResNet-50  & \Checkmark & 83.3M & 29.0        &   48.0     &    30.3     \\ 
            \textbf{FPN300/500}~\cite{lin2016feature}   &  ResNet-101  & \Checkmark & 121.2M &   29.4        &   48.8     &    30.6   \\ 
            \textbf{Mask RCNN+FPN300/500}~\cite{he2017mask}   &  ResNet-50  & \Checkmark & 84.4M &     29.9       &  49.0    &   31.3     \\ 
            \textbf{Mask RCNN+FPN300/500}~\cite{he2017mask}   &  ResNet-101  & \Checkmark & 122.4M &     30.2       &  \textbf{49.3}    &   31.7   \\  
\hline
		\end{tabular}
	}
\end{table*}

\vspace{-.8em}
\subsection{Results on MS COCO}
Finally we evaluate our DSOD on the MS COCO dataset~\cite{lin2014microsoft}. MS COCO contains 80k images for training, 40k for validation and 20k for testing ({\texttt {test-dev}} set).
Following~\cite{ren2015faster,li2016r}, we use the {\texttt {trainval}} set (train set + validation set) for training.
The batch size is also set as 128.
The initial learning rate is set to 0.1 for the first 80k iterations, then divided by 10 after every 60k iterations. The total number of training iterations is 320k.

Results are summarized in Tab.~\ref{COCO}. Our DSOD300 achieves 29.3\%/47.3\% on the {\texttt {test-dev}} set, which outperforms the baseline SSD300$^*$ with a large margin. Our result is comparable to the single-scale R-FCN, and is close to the R-FCN{\footnotesize{multi-sc}} which uses ResNet-101 as the pre-trained model.
Interestingly, we observe that our result with 0.5 IoU is lower than R-FCN, but our [0.5:0.95] result is better or comparable. This indicates that our predicted locations are more accurate than R-FCN under the larger overlap settings. It is reasonable that our small object detection precision is slightly lower than R-FCN since our input image size (300$\times$300) is much smaller than R-FCN's ($\sim$ 600$\times$1000). Even with this disadvantage, our large object detection precision is still much better than R-FCN. This further demonstrates the effectiveness of our approach. Fig.~\ref{examples} shows some qualitative detection examples on COCO with our DSOD300 model.

\vspace{-.8em}
\subsection{From MS COCO to PASCAL VOC}
Next, we investigate how the MS COCO dataset can further help with the detection performance on PASCAL VOC. We use the DSOD model trained on the COCO (without the ImageNet pre-trained model) to initialize the network weights. Then another DSOD is fine-tuned on PASCAL VOC datasets with small initial learning rate (0.001). This operation leads to 81.7\% mAP on PASCAL VOC 2007 and 79.3\% mAP on PASCAL VOC 2012, respectively. The extra data from the COCO set increases the mAP by 4.0\% on PASCAL VOC 2007 and 3.0\% on VOC 2012. The results verify that although our DSOD models are trained with fewer images, they have not overfitted to the PASCAL VOC datasets yet, and still have room to be boosted.

\vspace{-.8em}
\subsection{From DSOD to DSOD (v2)}
Compared with DSOD, DSOD v2 includes the extra DSS module to further enhance the supervision signal under the training from scratch scenario. The comparison results of DSOD and DSOD v2 are shown in Tab.~\ref{comparisons}. We can see that DSOD v2 improves the performance consistently on both PASCAL VOC and COCO datasets under different training sets. In DSOD v2, we also replace the pre-activation of BN~\cite{he2016identity} in DSOD with post-activation (replacing BN-ReLU-Conv with the Conv-BN-ReLU manner), as shown in Fig.~\ref{post_act}. We observe that this operation can improve the detection performance with about 0.6 \% mAP.

\begin{figure*}[]
	\centering
	\includegraphics[width=0.96\textwidth]{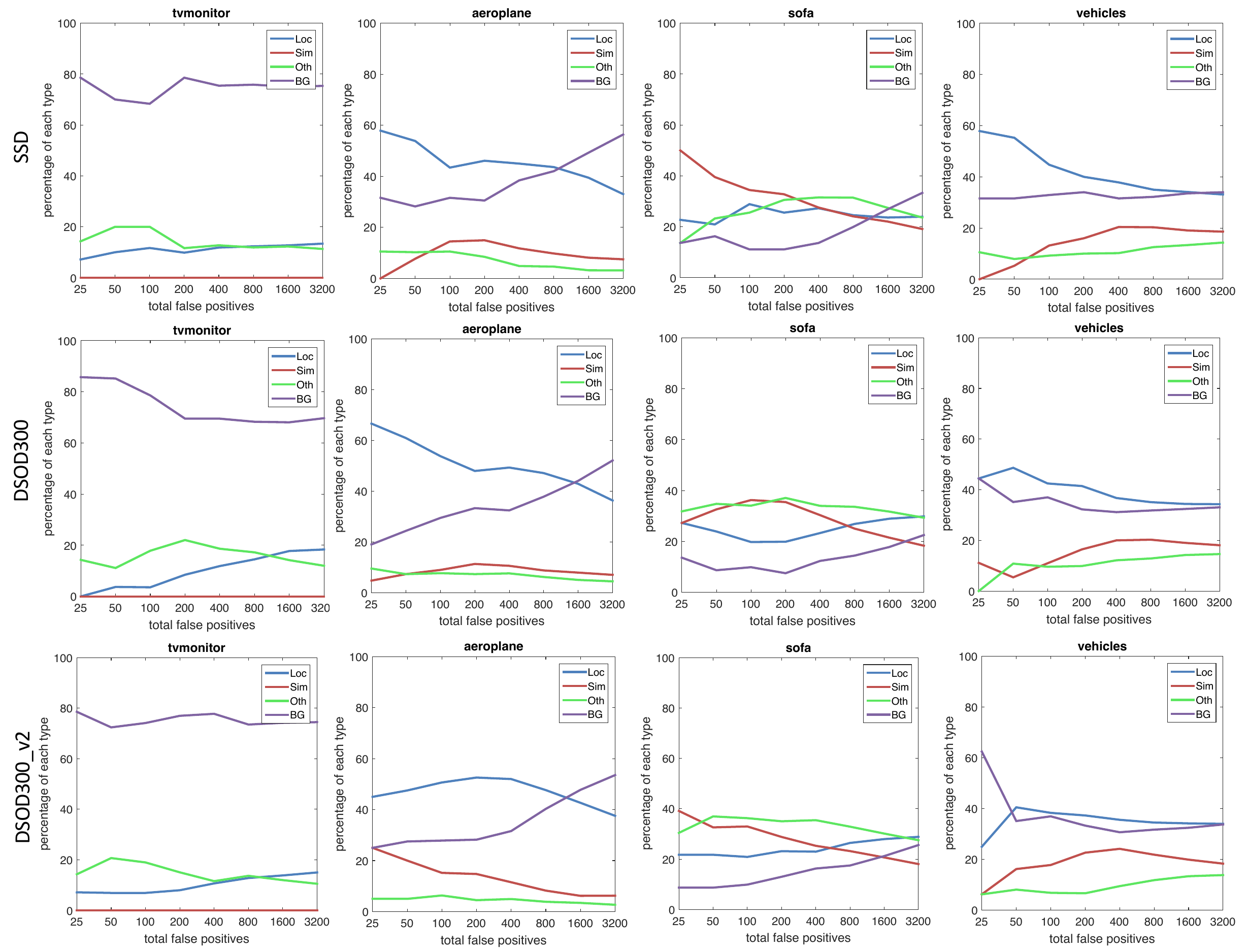}
	\vspace{-0.12in}
	\caption{Distributions and trendlines of top-ranked false positive (FP) types.
Each plot shows the evolving distribution (trendline) of FP types as more FP examples
are considered. Each FP is categorized into 1 of 4 types: Loc: poor localization (a detection with an IoU overlap with the correct class between 0.1 and 0.5); Sim: confusion with a similar category; Oth: confusion with a dissimilar object category; BG: a FP that fired on background. More details can be referred to~\cite{hoiem2012diagnosing}.}
	\label{distribution}
	\vspace{-0.12in}
\end{figure*}

\vspace{-.8em}
\subsection{Comparisons of State-of-the-art Two-Stage Detectors} \label{48}
In this section, we compare our results with the state-of-the-art two-stage detectors, including Faster RCNN + FPN and Mask RCNN + FPN. For fair comparisons, we resize the short side of inputs to 300 for these two-stage detectors. The whole comparisons are shown in Tab.~\ref{one_two}. We can observe that DSOD300 (29.3\% mAP) achieves comparable results with FPN300/500 (ResNet-101 backbone, 29.4\% mAP), while the \#params of DSOD (21.8M) is only about 1/6 compared to FPN300/500 with ResNet-101 (121.2M). The performance of our DSOD300* v2 (30.4\% mAP) is even slightly better than Mask RCNN + FPN300/500 with ResNet-101 (30.2\% mAP) while requiring only 1/3 of parameters (37.3M {\em vs.} 120.6M). The results show great advantages and potential of our proposed methods. 

\vspace{-.8em}
\subsection{Comparisons of Different Input Sizes} \label{49}
Intuitively, larger input images will bring better performance for object detection. We verify this by using different input resolutions with: 300, 360, 440, 512 and maintaining 4 images on each GPU during training (the total batch size is still 128). The results on PASCAL VOC are illustrated in Fig.~\ref{resolutions}. We can observe that larger input can obtain higher accuracy, which is consistent to our conjecture.

\begin{figure}[]
	\centering
	\includegraphics[width=0.38\textwidth]{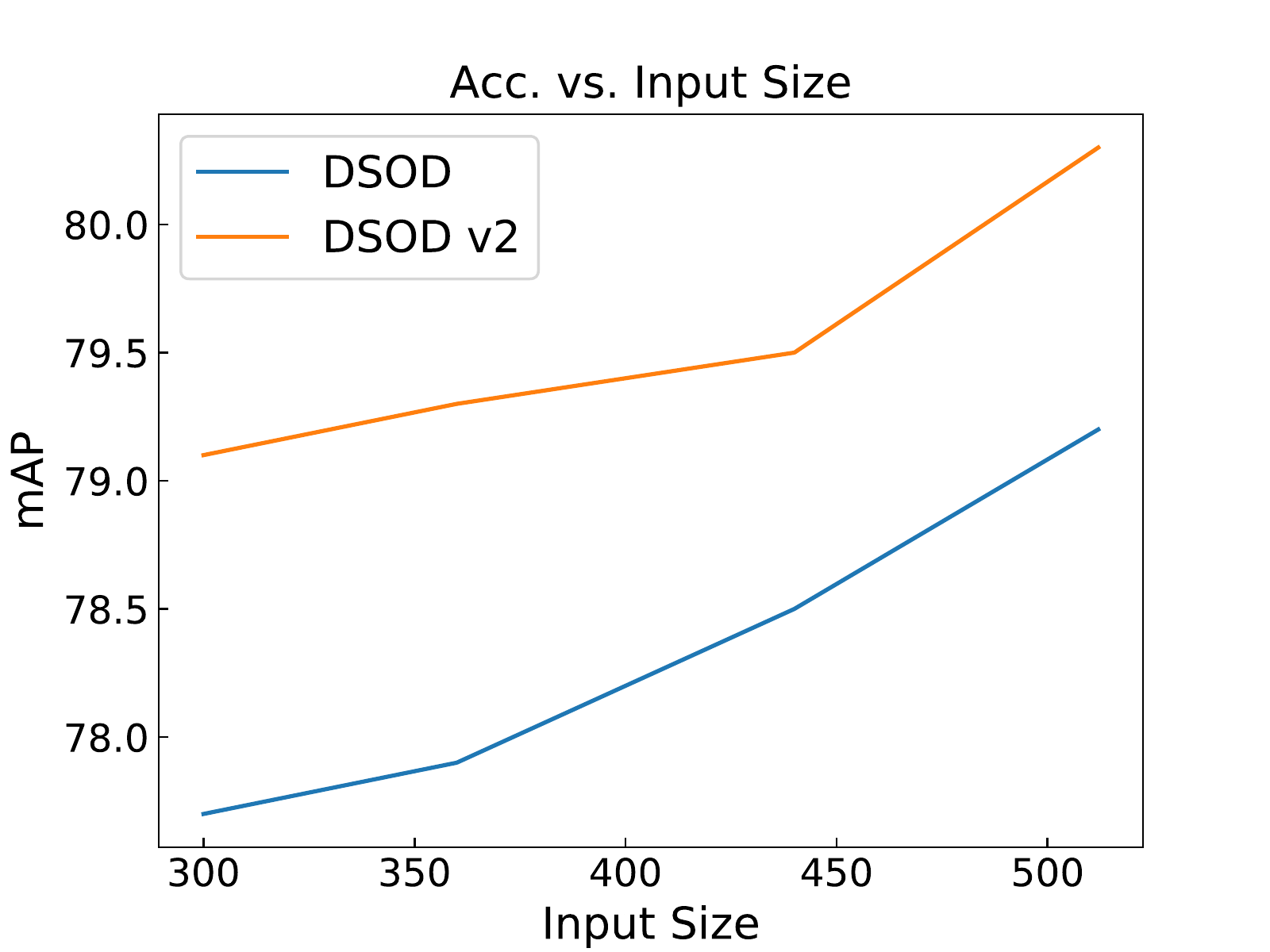}
	\vspace{-0.12in}
	\caption{Accuracy under different input sizes.}
	\label{resolutions}
	\vspace{-0.22in}
\end{figure}

\vspace{-.8em}
\subsection{Models and Results Analysis}
In order to reveal the failure reasons of our methods and the error differences between baseline SSD and our methods, we conduct experiments on the following two aspects of analysis, including: (1) the sensitivity to object characteristics, shown in Fig.~\ref{sensitivity}; (2) the distribution and trendline of top-ranked false positive (FP) types, as shown in Fig.~\ref{distribution}. We adopted the publicly available detection analysis tool from Hoiem et al.~\cite{hoiem2012diagnosing} for these illustrations. More explanation can be referred to the captions under these two figures.


\begin{figure*}[]
	\centering
	\includegraphics[width=0.70\textwidth]{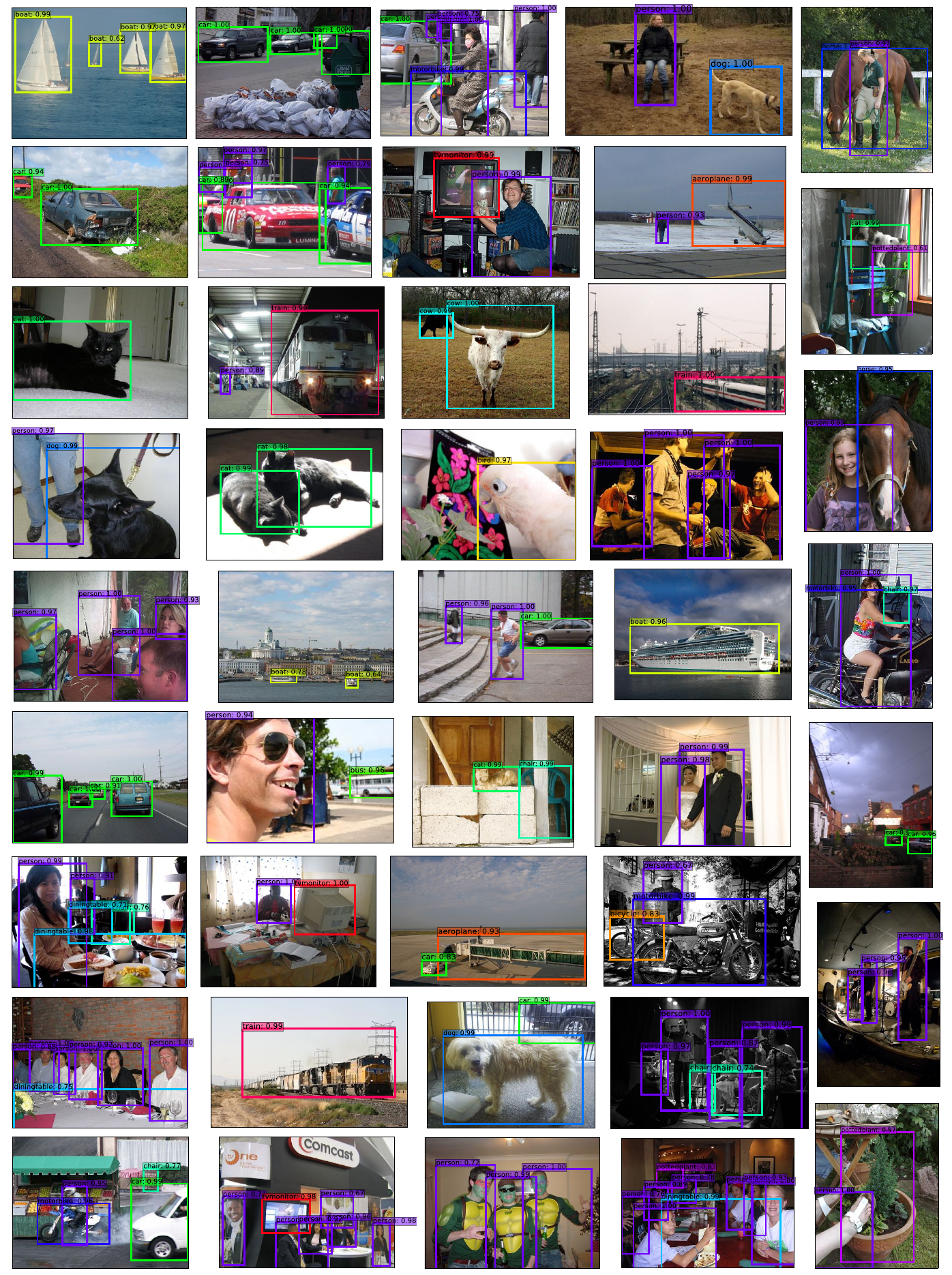}
	\vspace{-0.12in}
	\caption{More examples of object detection results on the PASCAL VOC 2012 {\em {test}} set using DSOD300. The training data is VOC 2007 {\em trainval}, VOC 2007 {\em test}, VOC 2012 {\em trainval} and MS COCO {\em {trainval}} (79.3\% mAP@0.5 on the {\em {test}} set). Each output box is associated with a category label and a softmax score in [0, 1]. A score threshold of 0.6 is used for displaying. For each image, one color corresponds to one object category in that image.}
	\label{voc12}
	\vspace{-0.12in}
\end{figure*}

\section{Discussion}\label{discussion}
\noindent{\textbf{Better Model Structure {\em{vs.}} More Training Data.}}
An emerging idea in the computer vision community is that object detection or other vision tasks might be solved with deeper and larger neural networks backed with massive training data like ImageNet~\cite{deng2009imagenet}.
Thus more and more large-scale datasets have been collected and released recently, such as the Open Images dataset~\cite{openimages},
which is 7.5x larger in the number of images and 6x larger of categories than that of ImageNet.
We definitely agree that, under modest assumptions that given boundless training data and unlimited computational power, deep neural networks should perform extremely well. However, our proposed approach and experimental results imply an alternative view to handle this problem: a better model structure might enable similar or better performance compared with complex models trained from large data.
Particularly, our DSOD is only trained with 16,551 images on VOC 2007, but achieves competitive or even better performance than those models trained with 1.2 million + 16,551 images.

In this premise, it is worthwhile rehashing the intuition that as datasets grow larger, training deep neural networks becomes more and more expensive. Thus a simple yet efficient approach becomes increasingly important. Despite its conceptual simplicity, our approach shows great potential under this setting.

\noindent{\textbf{Why Training from Scratch?}}
There are many successful cases that fine-tuning works well and achieves consistent improvement, especially in object detection areas. So why do we still need to train object detectors from scratch?
As aforementioned briefly, the critical importance of training from scratch has at least two aspects.
First, there may have big domain differences between the pre-trained and the target one.
For instance, most pre-trained models are learned on large-scale RGB dataset like ImageNet.
It is fairly difficult to transfer RGB models to depth images, multi-spectrum images, medical images, etc.
Some advanced domain adaptation techniques have been proposed and could mitigate this problem.
But what an amazing thing if we have a technique that can train object detector from scratch.
Second, fine-tuning restricts the design space of network structures for object detection. This is very critical for the deployment of applying deep neural networks to some resource-limited Internet-of-Things (IoT) scenario.

\noindent{\textbf{Model Compactness {\em{vs.}} Performance.}}
Model compactness (in terms of the number of parameters) and performance is an important trade-off for the applications of deep neural networks in actual detection scenarios.
Most CNN-based detection solutions require a huge memory space to store the massive parameters.
Therefore the models are usually unsuitable for low-end devices like mobile-phones and embedded electronics.
Thanks to the parameter-efficient dense connections, our model is much smaller than most competitive methods.
For instance, our smallest dense model (DS/64-64-16-1, with dense prediction layers) achieves 73.6\% mAP with only 5.9M parameters, which shows great potential for applications on low-end devices. Adopting network pruning methods~\cite{han2015deep,liu2017learning} to further reduce the parameters and speed up the inference process will be a good direction for CNN-based object detection, and will be investigated in the further.

\noindent{\textbf{How to Train Two-Stage Detectors from Scratch.}}
Some recent works~\cite{he2018rethinking,wu2018group} have observed that utilizing new techniques (e.g., Sync BN~\cite{peng2018megdet}, Group Norm~\cite{wu2018group}, Switchable Norm~\cite{luo2018differentiable}, etc.) and more training epochs could enable to train two-stage detectors from scratch.
We also did some preliminary experiments on PASCAL VOC 2007 dataset (limited training data) with two-stage detectors from scratch (use VGG16 as backbone network and with standard training budget). As shown in Tab.~\ref{two}, our results indicates that if replacing RoI Pool with RoI Align and adopting advanced normalization methods can enable to train two-stage detectors from scratch. 

\begin{table}
	\centering
	\caption{Comparison of performance (mAP) using different building designs in two-stage object detectors when training from scratch. The backbone network is VGG16~\cite{simonyan2014very}. All models are trained on VOC 07~\cite{everingham2010pascal} \texttt{trainval} set and tested on \texttt{test} set.}
	\resizebox{0.5\textwidth}{!}{%
		\label{ablation}
		\begin{tabular}{c|c|c|c|c}
			\hline
			BN     &  Sync\_BN  & RoI Pool   & RoI Align     &          mAP (\%)  \\ \hline \hline
			&                  & \Checkmark   &                    &        6.4  \\ \hline
			\Checkmark &                  & \Checkmark   &                    &        37.3 \\ \hline
			&                  &                      &\Checkmark  &        36.1  \\ \hline
			\Checkmark &                  &                      & \Checkmark &        48.7 \\ \hline
			&\Checkmark &                     & \Checkmark &          50.5 \\ \hline
		\end{tabular}
	}
	\vspace{-2ex}
	\label{two}
\end{table}

\section{Conclusion}\label{conclusion}
We have presented Deeply Supervised Object Detector (DSOD), a simple yet efficient framework for learning object detectors from scratch. Without using pre-trained models from ImageNet, DSOD demonstrates competitive performance to state-of-the-art detectors such as SSD, Faster R-CNN, R-FCN, FPN, Mask RCNN, etc. on the popular PASCAL VOC 2007, 2012 and MS COCO datasets,
meanwhile, with only 1/2, 1/4 and 1/10 parameters compared to SSD, R-FCN and Faster R-CNN, respectively.
Due to the learning from scratch property, DSOD has great potential on domain-different scenarios, such ad depth, medical, multi-spectral images, etc. Our future work will consider learning object detectors directly in these diverse domains,
as well as learning ultra efficient DSOD models to support resource-bounded devices.

\section*{Acknowledgements}
Yu-Gang Jiang and Xiangyang Xue were supported in part by National Key R\&D Program of China (No.2017YFC0803700), NSFC under Grant (No.61572138 \& No.U1611461) and STCSM Project under Grant No.16JC1420400.


\ifCLASSOPTIONcaptionsoff
  \newpage
\fi

\bibliographystyle{IEEEtran}
\bibliography{DSOD}

\begin{thebibliography}{10}
\providecommand{\url}[1]{#1}
\csname url@samestyle\endcsname
\providecommand{\newblock}{\relax}
\providecommand{\bibinfo}[2]{#2}
\providecommand{\BIBentrySTDinterwordspacing}{\spaceskip=0pt\relax}
\providecommand{\BIBentryALTinterwordstretchfactor}{4}
\providecommand{\BIBentryALTinterwordspacing}{\spaceskip=\fontdimen2\font plus
\BIBentryALTinterwordstretchfactor\fontdimen3\font minus
  \fontdimen4\font\relax}
\providecommand{\BIBforeignlanguage}[2]{{%
\expandafter\ifx\csname l@#1\endcsname\relax
\typeout{** WARNING: IEEEtran.bst: No hyphenation pattern has been}%
\typeout{** loaded for the language `#1'. Using the pattern for}%
\typeout{** the default language instead.}%
\else
\language=\csname l@#1\endcsname
\fi
#2}}
\providecommand{\BIBdecl}{\relax}
\BIBdecl

\bibitem{he2017mask}
K.~He, G.~Gkioxari, P.~Doll{\'a}r, and R.~Girshick, ``Mask r-cnn,'' in
  \emph{ICCV}, 2017.

\bibitem{lin2016feature}
T.-Y. Lin, P.~Doll{\'a}r, R.~Girshick, K.~He, B.~Hariharan, and S.~Belongie,
  ``Feature pyramid networks for object detection,'' in \emph{CVPR}, 2017.

\bibitem{deng2009imagenet}
J.~Deng, W.~Dong, R.~Socher, L.-J. Li \emph{et~al.}, ``Imagenet: A large-scale
  hierarchical image database,'' in \emph{CVPR}, 2009.

\bibitem{oquab2014learning}
M.~Oquab, L.~Bottou, I.~Laptev, and J.~Sivic, ``Learning and transferring
  mid-level image representations using convolutional neural networks,'' in
  \emph{CVPR}, 2014.

\bibitem{cui2018transfer}
W.~Cui, G.~Zheng, Z.~Shen, S.~Jiang, and W.~Wang, ``Transfer learning for
  sequences via learning to collocate,'' in \emph{International Conference on
  Learning Representations}, 2019.

\bibitem{girshick2014rich}
R.~Girshick, J.~Donahue, T.~Darrell, and J.~Malik, ``Rich feature hierarchies
  for accurate object detection and semantic segmentation,'' in \emph{CVPR},
  2014.

\bibitem{girshick2015fast}
R.~Girshick, ``Fast r-cnn,'' in \emph{ICCV}, 2015.

\bibitem{ren2015faster}
S.~Ren, K.~He, R.~Girshick, and J.~Sun, ``Faster r-cnn: Towards real-time
  object detection with region proposal networks,'' in \emph{NIPS}, 2015.

\bibitem{li2016r}
Y.~Li, K.~He, J.~Sun \emph{et~al.}, ``R-fcn: Object detection via region-based
  fully convolutional networks,'' in \emph{NIPS}, 2016.

\bibitem{liu2016ssd}
W.~Liu, D.~Anguelov, D.~Erhan \emph{et~al.}, ``Ssd: Single shot multibox
  detector,'' in \emph{ECCV}, 2016.

\bibitem{redmon2016you}
J.~Redmon, S.~Divvala, R.~Girshick, and A.~Farhadi, ``You only look once:
  Unified, real-time object detection,'' in \emph{CVPR}, 2016.

\bibitem{lin2017focal}
T.-Y. Lin, P.~Goyal, R.~Girshick, K.~He, and P.~Doll{\'a}r, ``Focal loss for
  dense object detection,'' in \emph{Proceedings of the IEEE international
  conference on computer vision}, 2017, pp. 2980--2988.

\bibitem{kong2016hypernet}
T.~Kong, A.~Yao, Y.~Chen, and F.~Sun, ``Hypernet: Towards accurate region
  proposal generation and joint object detection,'' in \emph{Proceedings of the
  IEEE conference on computer vision and pattern recognition}, 2016, pp.
  845--853.

\bibitem{bell2016inside}
S.~Bell, C.~Lawrence~Zitnick \emph{et~al.}, ``Inside-outside net: Detecting
  objects in context with skip pooling and recurrent neural networks,'' in
  \emph{CVPR}, 2016.

\bibitem{kong2017ron}
T.~Kong, F.~Sun, A.~Yao, H.~Liu, M.~Lu, and Y.~Chen, ``Ron: Reverse connection
  with objectness prior networks for object detection,'' in \emph{Proceedings
  of the IEEE Conference on Computer Vision and Pattern Recognition}, 2017, pp.
  5936--5944.

\bibitem{peng2018megdet}
C.~Peng, T.~Xiao, Z.~Li, Y.~Jiang, X.~Zhang, K.~Jia, G.~Yu, and J.~Sun,
  ``Megdet: A large mini-batch object detector,'' in \emph{Proceedings of the
  IEEE Conference on Computer Vision and Pattern Recognition}, 2018, pp.
  6181--6189.

\bibitem{singh2018analysis}
B.~Singh and L.~S. Davis, ``An analysis of scale invariance in object detection
  snip,'' in \emph{Proceedings of the IEEE Conference on Computer Vision and
  Pattern Recognition}, 2018, pp. 3578--3587.

\bibitem{hu2018relation}
H.~Hu, J.~Gu, Z.~Zhang, J.~Dai, and Y.~Wei, ``Relation networks for object
  detection,'' in \emph{Proceedings of the IEEE Conference on Computer Vision
  and Pattern Recognition}, 2018, pp. 3588--3597.

\bibitem{cai2018cascade}
Z.~Cai and N.~Vasconcelos, ``Cascade r-cnn: Delving into high quality object
  detection,'' in \emph{Proceedings of the IEEE Conference on Computer Vision
  and Pattern Recognition}, 2018, pp. 6154--6162.

\bibitem{bosquetstdnet}
B.~Bosquet, M.~Mucientes, and V.~M. Brea, ``Stdnet: A convnet for small target
  detection,'' in \emph{BMVC}, 2018.

\bibitem{xu2018deep}
H.~Xu, X.~Lv, X.~Wang, Z.~Ren, N.~Bodla, and R.~Chellappa, ``Deep regionlets
  for object detection,'' in \emph{Proceedings of the European Conference on
  Computer Vision (ECCV)}, 2018, pp. 798--814.

\bibitem{wang2018pelee}
R.~J. Wang, X.~Li, and C.~X. Ling, ``Pelee: A real-time object detection system
  on mobile devices,'' in \emph{Advances in Neural Information Processing
  Systems}, 2018, pp. 1963--1972.

\bibitem{long2015fully}
J.~Long, E.~Shelhamer, and T.~Darrell, ``Fully convolutional networks for
  semantic segmentation,'' in \emph{CVPR}, 2015.

\bibitem{hariharan2015hypercolumns}
B.~Hariharan, P.~Arbel{\'a}ez, R.~Girshick, and J.~Malik, ``Hypercolumns for
  object segmentation and fine-grained localization,'' in \emph{CVPR}, 2015.

\bibitem{chen2014semantic}
L.-C. Chen, G.~Papandreou, I.~Kokkinos \emph{et~al.}, ``Semantic image
  segmentation with deep convolutional nets and fully connected crfs,'' in
  \emph{ICLR}, 2015.

\bibitem{yu2015multi}
F.~Yu and V.~Koltun, ``Multi-scale context aggregation by dilated
  convolutions,'' in \emph{ICLR}, 2016.

\bibitem{zhang2014part}
N.~Zhang, J.~Donahue, R.~Girshick, and T.~Darrell, ``Part-based r-cnns for
  fine-grained category detection,'' in \emph{ECCV}, 2014.

\bibitem{lin2015bilinear}
T.-Y. Lin, A.~RoyChowdhury, and S.~Maji, ``Bilinear cnn models for fine-grained
  visual recognition,'' in \emph{ICCV}, 2015.

\bibitem{wang2015multiple}
D.~Wang, Z.~Shen, J.~Shao, W.~Zhang, X.~Xue, and Z.~Zhang, ``Multiple
  granularity descriptors for fine-grained categorization,'' in \emph{ICCV},
  2015.

\bibitem{krause2015fine}
J.~Krause, H.~Jin, J.~Yang, and L.~Fei-Fei, ``Fine-grained recognition without
  part annotations,'' in \emph{CVPR}, 2015.

\bibitem{xu2015show}
K.~Xu, J.~Ba, R.~Kiros, K.~Cho, A.~Courville, R.~Salakhudinov, R.~Zemel, and
  Y.~Bengio, ``Show, attend and tell: Neural image caption generation with
  visual attention,'' in \emph{ICML}, 2015.

\bibitem{donahue2015long}
J.~Donahue, L.~Anne~Hendricks, S.~Guadarrama, M.~Rohrbach, S.~Venugopalan,
  K.~Saenko, and T.~Darrell, ``Long-term recurrent convolutional networks for
  visual recognition and description,'' in \emph{CVPR}, 2015.

\bibitem{fang2015captions}
H.~Fang, S.~Gupta, and et~al, ``From captions to visual concepts and back,'' in
  \emph{CVPR}, 2015.

\bibitem{shen2017weakly}
Z.~Shen, J.~Li, Z.~Su, M.~Li, Y.~Chen, Y.-G. Jiang, and X.~Xue, ``Weakly
  supervised dense video captioning,'' in \emph{CVPR}, 2017.

\bibitem{krishna2017dense}
R.~Krishna, K.~Hata, F.~Ren, L.~Fei-Fei, and J.~C. Niebles, ``Dense-captioning
  events in videos,'' in \emph{ICCV}, 2017.

\bibitem{johnson2016densecap}
J.~Johnson, A.~Karpathy, and L.~Fei-Fei, ``Densecap: Fully convolutional
  localization networks for dense captioning,'' in \emph{CVPR}, 2016.

\bibitem{gupta2016cross}
S.~Gupta, J.~Hoffman, and J.~Malik, ``Cross modal distillation for supervision
  transfer,'' in \emph{CVPR}, 2016.

\bibitem{lee2015deeply}
C.-Y. Lee, S.~Xie, P.~W. Gallagher \emph{et~al.}, ``Deeply-supervised nets.''
  in \emph{AISTATS}, 2015.

\bibitem{xie2015hed}
S.~Xie and Z.~Tu, ``Holistically-nested edge detection,'' in \emph{ICCV}, 2015.

\bibitem{huang2016densely}
G.~Huang, Z.~Liu, K.~Q. Weinberger, and L.~van~der Maaten, ``Densely connected
  convolutional networks,'' in \emph{CVPR}, 2017.

\bibitem{Shen2017DSOD}
Z.~Shen, Z.~Liu, J.~Li, Y.-G. Jiang, Y.~Chen, and X.~Xue, ``Dsod: Learning
  deeply supervised object detectors from scratch,'' in \emph{ICCV}, 2017.

\bibitem{shen2017learning}
Z.~Shen, H.~Shi, R.~Feris, L.~Cao, S.~Yan, D.~Liu, X.~Wang, X.~Xue, and T.~S.
  Huang, ``Learning object detectors from scratch with gated recurrent feature
  pyramids,'' \emph{arXiv preprint arXiv:1712.00886}, 2017.

\bibitem{li2018tiny}
Y.~Li, J.~Li, W.~Lin, and J.~Li, ``Tiny-dsod: Lightweight object detection for
  resource-restricted usages,'' \emph{arXiv preprint arXiv:1807.11013}, 2018.

\bibitem{uijlings2013selective}
J.~R. Uijlings, K.~E. Van De~Sande, T.~Gevers \emph{et~al.}, ``Selective search
  for object recognition,'' \emph{IJCV}, 2013.

\bibitem{zhou2018scale}
P.~Zhou, B.~Ni, C.~Geng, J.~Hu, and Y.~Xu, ``Scale-transferrable object
  detection,'' in \emph{Proceedings of the IEEE Conference on Computer Vision
  and Pattern Recognition}, 2018, pp. 528--537.

\bibitem{zhang2018single}
S.~Zhang, L.~Wen, X.~Bian, Z.~Lei, and S.~Z. Li, ``Single-shot refinement
  neural network for object detection,'' in \emph{Proceedings of the IEEE
  Conference on Computer Vision and Pattern Recognition}, 2018, pp. 4203--4212.

\bibitem{liu2018receptive}
S.~Liu, D.~Huang \emph{et~al.}, ``Receptive field block net for accurate and
  fast object detection,'' in \emph{Proceedings of the European Conference on
  Computer Vision (ECCV)}, 2018, pp. 385--400.

\bibitem{law2018cornernet}
H.~Law and J.~Deng, ``Cornernet: Detecting objects as paired keypoints,'' in
  \emph{Proceedings of the European Conference on Computer Vision (ECCV)},
  2018, pp. 734--750.

\bibitem{zhou2019bottom}
X.~Zhou, J.~Zhuo, and P.~Kr{\"a}henb{\"u}hl, ``Bottom-up object detection by
  grouping extreme and center points,'' \emph{arXiv preprint arXiv:1901.08043},
  2019.

\bibitem{krizhevsky2012imagenet}
A.~Krizhevsky, I.~Sutskever, and G.~Hinton, ``Imagenet classification with deep
  convolutional neural networks,'' in \emph{NIPS}, 2012.

\bibitem{simonyan2014very}
K.~Simonyan and A.~Zisserman, ``Very deep convolutional networks for
  large-scale image recognition,'' in \emph{ICLR}, 2015.

\bibitem{szegedy2015going}
C.~Szegedy, W.~Liu, Y.~Jia, P.~Sermanet \emph{et~al.}, ``Going deeper with
  convolutions,'' in \emph{CVPR}, 2015.

\bibitem{he2016deep}
K.~He, X.~Zhang, S.~Ren, and J.~Sun, ``Deep residual learning for image
  recognition,'' in \emph{CVPR}, 2016.

\bibitem{srivastava2014dropout}
N.~Srivastava, G.~E. Hinton, A.~Krizhevsky \emph{et~al.}, ``Dropout: a simple
  way to prevent neural networks from overfitting.'' \emph{JMLR}, 2014.

\bibitem{ioffe2015batch}
S.~Ioffe and C.~Szegedy, ``Batch normalization: Accelerating deep network
  training by reducing internal covariate shift,'' \emph{arXiv preprint
  arXiv:1502.03167}, 2015.

\bibitem{redmon2016yolo9000}
J.~Redmon and A.~Farhadi, ``Yolo9000: Better, faster, stronger,'' in
  \emph{CVPR}, 2017.

\bibitem{redmon2018yolov3}
------, ``Yolov3: An incremental improvement,'' \emph{arXiv preprint
  arXiv:1804.02767}, 2018.

\bibitem{kim2016pvanet}
K.-H. Kim, S.~Hong, B.~Roh \emph{et~al.}, ``Pvanet: Deep but lightweight neural
  networks for real-time object detection,'' \emph{arXiv preprint
  arXiv:1608.08021}, 2016.

\bibitem{huang2016speed}
J.~Huang, V.~Rathod, C.~Sun \emph{et~al.}, ``Speed/accuracy trade-offs for
  modern convolutional object detectors,'' in \emph{CVPR}, 2017.

\bibitem{szegedy2016inception}
C.~Szegedy, S.~Ioffe, V.~Vanhoucke, and A.~Alemi, ``Inception-v4,
  inception-resnet and the impact of residual connections on learning,'' in
  \emph{ICLR workshop}, 2016.

\bibitem{jegou2016one}
S.~J{\'e}gou, M.~Drozdzal, D.~Vazquez \emph{et~al.}, ``The one hundred layers
  tiramisu: Fully convolutional densenets for semantic segmentation,''
  \emph{arXiv preprint arXiv:1611.09326}, 2016.

\bibitem{peng2017megdet}
C.~Peng, T.~Xiao, Z.~Li, Y.~Jiang, X.~Zhang, K.~Jia, G.~Yu, and J.~Sun,
  ``Megdet: A large mini-batch object detector,'' \emph{arXiv preprint
  arXiv:1711.07240}, 2017.

\bibitem{wu2018group}
Y.~Wu and K.~He, ``Group normalization,'' \emph{arXiv preprint
  arXiv:1803.08494}, 2018.

\bibitem{sun2015deepid3}
Y.~Sun, D.~Liang, X.~Wang, and X.~Tang, ``Deepid3: Face recognition with very
  deep neural networks,'' \emph{arXiv preprint arXiv:1502.00873}, 2015.

\bibitem{szegedy2016rethinking}
C.~Szegedy, V.~Vanhoucke, S.~Ioffe \emph{et~al.}, ``Rethinking the inception
  architecture for computer vision,'' in \emph{CVPR}, 2016.

\bibitem{jia2014caffe}
Y.~Jia, E.~Shelhamer, J.~Donahue \emph{et~al.}, ``Caffe: Convolutional
  architecture for fast feature embedding,'' in \emph{ACM MM}, 2014.

\bibitem{liu2015parsenet}
W.~Liu, A.~Rabinovich, and A.~C. Berg, ``Parsenet: Looking wider to see
  better,'' \emph{arXiv preprint arXiv:1506.04579}, 2015.

\bibitem{glorot2010understanding}
X.~Glorot and Y.~Bengio, ``Understanding the difficulty of training deep
  feedforward neural networks.'' in \emph{AISTATS}, 2010.

\bibitem{lin2014microsoft}
T.-Y. Lin, M.~Maire, S.~Belongie \emph{et~al.}, ``Microsoft coco: Common
  objects in context,'' in \emph{ECCV}, 2014.

\bibitem{he2016identity}
K.~He, X.~Zhang, S.~Ren, and J.~Sun, ``Identity mappings in deep residual
  networks,'' in \emph{ECCV}, 2016.

\bibitem{hoiem2012diagnosing}
D.~Hoiem, Y.~Chodpathumwan, and Q.~Dai, ``Diagnosing error in object
  detectors,'' in \emph{ECCV}, 2012.

\bibitem{openimages}
I.~Krasin, T.~Duerig, N.~Alldrin, A.~Veit \emph{et~al.}, ``Openimages: A public
  dataset for large-scale multi-label and multi-class image classification.''
  \emph{https://github.com/openimages}, 2016.

\bibitem{han2015deep}
S.~Han, H.~Mao, and W.~J. Dally, ``Deep compression: Compressing deep neural
  networks with pruning, trained quantization and huffman coding,'' in
  \emph{ICLR}, 2016.

\bibitem{liu2017learning}
Z.~Liu, J.~Li, Z.~Shen, G.~Huang, S.~Yan, and C.~Zhang, ``Learning efficient
  convolutional networks through network slimming,'' in \emph{ICCV}, 2017.

\bibitem{he2018rethinking}
K.~He, R.~Girshick, and P.~Doll{\'a}r, ``Rethinking imagenet pre-training,''
  \emph{arXiv preprint arXiv:1811.08883}, 2018.

\bibitem{luo2018differentiable}
P.~Luo, J.~Ren, and Z.~Peng, ``Differentiable learning-to-normalize via
  switchable normalization,'' \emph{arXiv preprint arXiv:1806.10779}, 2018.

\bibitem{everingham2010pascal}
M.~Everingham, L.~Van~Gool, C.~K. Williams, J.~Winn, and A.~Zisserman, ``The
  pascal visual object classes (voc) challenge,'' \emph{International journal
  of computer vision}, vol.~88, no.~2, pp. 303--338, 2010.

\end{thebibliography}

\end{document}